\documentclass[letterpaper, 10 pt, journal, twoside]{IEEEtran}
\usepackage{amssymb, amsthm, amsmath,  cite, color}
\usepackage{booktabs}
\usepackage{float}
\usepackage{graphicx}
\usepackage{hyperref}
\usepackage{import}
\usepackage{mathtools}
\usepackage{multirow}
\usepackage{outline}
\usepackage{paralist}
\usepackage{siunitx}
\usepackage{stmaryrd}
\usepackage{systeme}
\usepackage{siunitx}
\usepackage{url}
\usepackage{balance}
\usepackage{tikz}
\usepackage{booktabs}
\usepackage{subcaption}
\usepackage{algorithm}
\usepackage{algorithmic}
\usepackage{tablefootnote}
\usepackage[font={small}]{caption}
\usepackage{bm}

\usepackage{enumitem}
\usepackage{tabularx}

\newtheorem{remark}{Remark}

\DeclareMathOperator*{\argmin}{argmin}

\begin{document}
\title{\LARGE \bf Velocity Obstacle for Polytopic Collision Avoidance \\ for Distributed Multi-robot Systems}
% Velocity Obstacle for Polytopes with Application to Collision Avoidance between Distributed Multi-robot System

\author{
\IEEEauthorblockN{Jihao Huang$^{1*}$, Jun Zeng$^{2*}$, Xuemin Chi$^{1}$, Koushil Sreenath$^{2}$, Zhitao Liu$^{1\dagger}$, Hongye Su$^{1}$}
\thanks{$^*$ Authors have contributed equally, $^\dagger$ Corresponding author.}
\thanks{This work was supported in part by National Key R\&D Program of China (Grant NO. 2021YFB3301000); National Natural Science Foundation of China (NSFC:62173297), Zhejiang Key R\&D Program (Grant NO. 2022C01035), Fundamental Research Funds for the Central Universities (NO.226-2022-00086).

\IEEEauthorblockA{$^1$ State Key Laboratory of Industrial Control Technology, Institute of Cyber-Systems and Control, Zhejiang University, Hangzhou, China {\tt\small  \{jihaoh, chixuemin, ztliu, hysu\}@zju.edu.cn}.} 

\IEEEauthorblockA{$^2$ Hybrid Robotics Group at the Department of Mechanical Engineering, UC Berkeley, USA {\tt\small  \{zengjunsjtu, koushils\}@berkeley.edu}.}

\IEEEauthorblockA{Simulation results are shown in the video \url{https://youtu.be/YT9aObT2VAo}.}

\IEEEauthorblockA{Codes are release in \url{https://github.com/HybridRobotics/vo-polytope}.}
}
}

\maketitle

%%%%%%%%%%%%
% abstract %
%%%%%%%%%%%%
\begin{abstract}
Obstacle avoidance for multi-robot navigation with polytopic shapes is challenging.
Existing works simplify the system dynamics or consider it as a convex or non-convex optimization problem with positive distance constraints between robots, which limits real-time performance and scalability.
Additionally, generating collision-free behavior for polytopic-shaped robots is harder due to implicit and non-differentiable distance functions between polytopes.
In this paper, we extend the concept of velocity obstacle (VO) principle for polytopic-shaped robots and propose a novel approach to construct the VO in the function of vertex coordinates and other robot's states.
Compared with existing work about obstacle avoidance between polytopic-shaped robots, our approach is much more computationally efficient as the proposed approach for construction of VO between polytopes is optimization-free. 
Based on VO representation for polytopic shapes, we later propose a navigation approach for distributed multi-robot systems.
We validate our proposed VO representation and navigation approach in multiple challenging scenarios including large-scale randomized tests, and our approach outperforms the state of art in many evaluation metrics, including completion rate, deadlock rate, and the average travel distance.
\end{abstract}

%%%%%%%%%%%%
% sections %
%%%%%%%%%%%%
\section{Introduction} 
\label{sec:intro}
\subsection{Motivation}
\begin{figure} 
    \centering
    \includegraphics[width=0.6\linewidth]{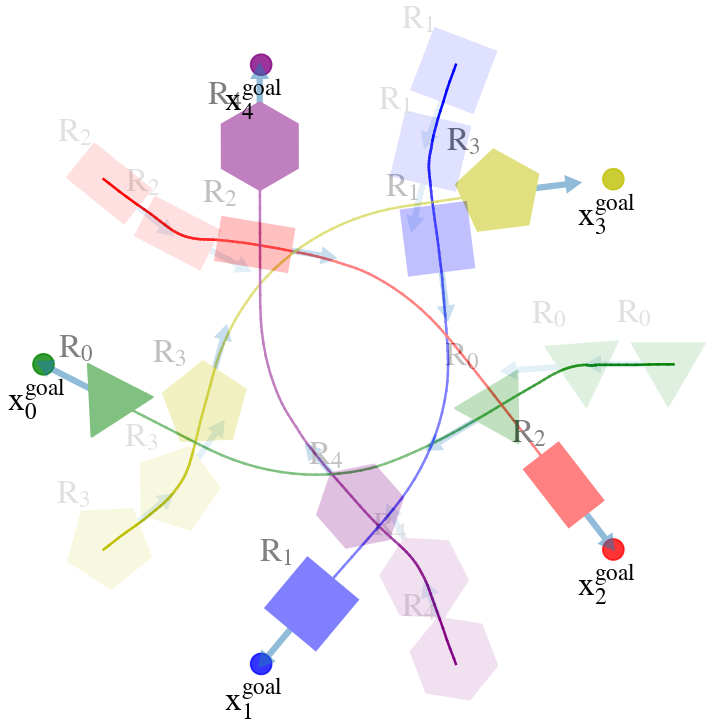}
    \caption{Snapshot of the distributed multi-robot navigation with polytopic shapes using our proposed approach. Each robot $\text{R}_i$ has its own destination $\mathbf{x}_{i}^{\text{goal}}$, and each robot could decide independently to move towards the destination while avoiding collisions. In this figure, we demonstrate each robot position at four different ticks with lighter shades of a color indicating robot's position in the past.}
    \hspace{-0.5cm}
    \label{fig:cover}
\end{figure}

\IEEEPARstart{T}{he} multi-robot navigation is a challenging task~\cite{chung2018survey, pian2022distributed} that needs to be solved for various applications such as warehouse delivery and search and rescue operations~\cite{queralta2020collaborative}.
The critical challenge of navigation tasks for the multi-robot system is to achieve real-time obstacle avoidance while navigating each robot to its respective destination, as shown in Fig.~\ref{fig:cover}.
Existing approaches are not accessible to be deployed for polytopic multi-robot systems, as they usually consider the collision avoidance between polytopes as convex or even non-convex optimizations, whose computational complexity increases dramatically with the number of robots.
Velocity obstacle\cite{fiorini1998motion}, commonly abbreviated VO, is the set of all velocities of a robot that will result in a collision with an obstacle at some moment in time, assuming that the obstacle maintains its current velocity.
In this paper, we propose a novel approach in the field of VO to achieve distributed multi-robot navigation with polytopic shapes, which could be deployed in real-time.
\subsection{Related Work}

\subsubsection{Collision Avoidance Between Polytopes}
Collision avoidance between polytopes is the crucial approach to achieve non-conservative collision avoidance for polytopic-shaped robots.
The polytopic shapes could be approximated into hyper-ellipses, however, this conservative over-approximation may lead to deadlock maneuvers, shown in~\cite{ziegler2010fast}.
Recently, optimization-based approaches have become popular in this field, where the obstacle avoidance criteria between polytopes are added as constraints.
However, the design of obstacle avoidance constraint is challenging since the distance function between two polytopes is also an optimization problem --- it is implicit and does not hold an analytical expression~\cite{grossmann2002review}.
Authors in~\cite{li2015unified} propose an optimization-based approach that keeps all the vertices of the controlled target outside all the other obstacles to achieve collision avoidance between rectangle-shaped robots, but this approach is neither applicable for the online calculation nor for the general polytopic shapes, especially in a densely packed environment.
To deal with obstacle avoidance between general polytopic shapes, mixed-integer programming~\cite{deits2015efficient} applies well for collision avoidance between two polytopic-shaped robots with linear dynamics, but it cannot be deployed as a real-time controller or trajectory planner for general nonlinear dynamical systems due to the added complexity from the nonlinear mixed-integer optimization problem.
A duality-based approach~\cite{zhang2018autonomous, zhang2020optimization} reformulates the implicit constraints of collision avoidance into smooth, differentiable constraints, which optimizes the computational complexity to some extent, but it is unsuitable for online planning of nonlinear systems.
This methodology has been extended into a distributed multi-robot system~\cite{firoozi2020distributed} which achieves real-time path planning for nonlinear systems based on a bi-level optimization scheme.
The dual approach has also been generalized for general nonlinear continuous-time~\cite{thirugnanam2021duality} and discrete-time~\cite{thirugnanam2022safety} systems to achieve real-time collision avoidance through convex~\cite{thirugnanam2021duality} or non-convex~\cite{thirugnanam2022safety} optimizations with control barrier functions~\cite{glotfelter2017nonsmooth, zeng2021safety}, but not in a distributed manner.
Notice that all the optimization-based approaches mentioned above~\cite{zhang2018autonomous, zhang2020optimization, firoozi2020distributed, thirugnanam2021duality, thirugnanam2022safety} usually suffer from a dramatic complexity increase with the number of the robots, which becomes challenging when deployed in real-time for a distributed multi-robot system.
In this paper we propose an optimization-free approach to address the above issues.

\subsubsection{Velocity Obstacle for Distributed Multi-Robot Navigation}
\label{subsub:dis}
Existing approaches for multi-robot navigation can be classified into centralized and distributed ones~\cite{asiain2020navigation}.
In contrast with centralized approaches, in distributed approaches each robot is an independent decision maker, and it could act independently according to the sensor information obtained from its onboard sensors~\cite{raibail2022decentralized}.
Since distributed approaches do not require global centralized communication between robots, it becomes suitable for a large-scale multi-robot system to achieve navigation tasks with limited computational resources.
One of its significant challenges is realizing reliable collision avoidance with limited sensor information and calculating optimal velocities with high efficiency and guaranteed safety.

Velocity obstacle (VO) and its variants have been widely used to realize local collision avoidance and navigation of the distributed multi-robot system in an open shared environment with static and dynamic obstacles.
VO~\cite{fiorini1998motion} is the set of all velocities of a robot that will result in a collision with an obstacle at some moment in time, assuming that the obstacle maintains its current velocity.
So each robot of the distributed multi-robot system could select a velocity which is outside of any VO induced by the obstacles or other robots and close to the preferred velocity in each time step to achieve collision avoidance and navigation independently.
Some variants of VO~\cite{van2008reciprocal, berg2011reciprocal, snape2011hybrid, alonso2012reciprocal, alonso2013optimal} make some improvements to reduce the unnecessary oscillations when using the VO-based algorithm for distributed multi-robot navigation, for more details readers can refer to Sec.~\ref{sec:vo_variants}.
% However, each robot has the reactive nature~\cite{van2008reciprocal} which enables it to independently adapt its velocity to avoid collision with other robots and obstacles, but VO disregards it.
% This could lead to unnecessary oscillations when using the VO-based algorithm for distributed multi-robot navigation.
% Some variants of VO explicitly consider this issue and make specific improvements to deal with the unnecessary oscillations, such as Reciprocal Velocity Obstacle (RVO)\cite{van2008reciprocal}, Hybrid Reciprocal Velocity Obstacle (HRVO)~\cite{snape2011hybrid}.
Some other works~\cite{long2018towards, li2019reciprocal, han2022reinforcement} also take full advantage of VO and combine it with reinforcement learning (RL) to learn a reliable collision avoidance policy for the distributed multi-robot navigation.

To summarize, VO and its variants are applicable for real-time navigation in distributed multi-robot systems, but current approaches are limited to circular robots and cannot be generalized to non-conservative avoidance of polytopic-shaped robots.
This paper proposes an extension of the VO principle to polytopic-shaped robots and obstacles and uses it to solve a distributed multi-robot navigation task.

\subsection{Contributions}
% contributions
In this paper, we mainly focus on the problem of distributed multi-robot navigation with polytopic-shaped robots and obstacles. 
The contributions of this paper are as follows:
\begin{itemize}
    \item We generalize the concept of VO to the polytopic-shaped robots and obstacles and propose a computationally efficient approach to construct the VO as a function of vertex coordinates and other robot's states.
    \item We propose a VO-based navigation approach for distributed multi-robot systems with polytopic shapes and validate it in many challenging scenarios, with multiple static and dynamic obstacles.
    \item We show that our proposed approach outperforms the state of the art in terms of completion rate, deadlock rate and the average travel distance through large-scale randomized tests.
\end{itemize}
\subsection{Organization}
% organize manners
The rest of this paper is organized as follows:
We formally define the problem of distributed multi-robot navigation with polytopic shapes, review the concept of VO and some variants of VO and describe the obstacle avoidance problem between polytopes in Sec.~\ref{sec:pre}.
In Sec.~\ref{sec:multi}, we describe how to construct the VO for polytopic-shaped robots and demonstrate how to achieve the distributed multi-robot navigation with polytopic shapes using VO.
We demonstrate our approach and present the simulation results in Sec.~\ref{sec:exper}.
Sec.~\ref{sec:con} concludes the paper.
\section{Problem Statement \& Preliminaries}
\label{sec:pre}
In this section, we firstly define the problem of distributed multi-robot navigation with polytopic shapes in Sec.~\ref{sec: problem statement}. Then, we present some preliminaries of velocity obstacle (VO), review two variants of VO (RVO and HRVO) and describe the obstacle avoidance problem between polytopes in Sec.~\ref{sec:vo_define}, ~\ref{sec:vo_variants} and ~\ref{sec:obstacle_avoidance_polytope}, respectively.
\subsection{Problem Statement}
\label{sec: problem statement}
% Define the problem
Assume there are a set of $N$ robots sharing an open space with a set of $M$ static and dynamic non-reactive obstacles, i.e., their velocities are not adapted to avoid collision with the other robots.
In this paper, we assume all robots and obstacles are polytopic-shaped.
For notations, subscripts $i$ and $j$ are applied to distinguish robots and obstacles, where each robot is represented by $\text{R}_i \in \mathbb{N}=\{\text{R}_0, \text{R}_1, \dots, \text{R}_{N-1}\}$ and each obstacle is represented by $\text{O}_j \in \mathbb{O}=\{\text{O}_0, \text{O}_1, \dots, \text{O}_{M-1}\}$,
and subscript $k$ is used to distinguish each vertex of a robot or an obstacle, i.e., $\text{v}_k$ represents the $k$th vertex.
For each robot in a distributed multi-robot system, there are external and internal states and only the external ones could be observed by other robots through the onboard sensors.
The external states of robot $\text{R}_i$ include the positions of all vertices $\mathbf{p}_{\text{R}_i}^{\text{v}_k}$, the current position $\mathbf{x}_{\text{R}_i}$ and the current velocity $\mathbf{v}_{\text{R}_i}$.
The robot's current position $\mathbf{x}_{\text{R}_i}$ is a function of the position of its vertices, where $\mathbf{x}_{\text{R}_i} = \frac{1}{K_{\text{R}_i}} \sum_{k=1}^{K_{\text{R}_i}} \mathbf{p}_{\text{R}_i}^{\text{v}_k}$ with the assumption that $\text{R}_i$ has $K_{\text{R}_i}$ vertices.
The current velocity $\mathbf{v}_{\text{R}_i}$ represents the time-derivative of the current position $\mathbf{x}_{\text{R}_i}$.
The internal states include the goal position $\mathbf{x}_{i}^{\text{goal}}$ and the maximum speed $v_{\text{R}_i, \text{max}}$.
The states of the obstacle $\text{O}_j$ include the current position $\mathbf{x}_{\text{O}_j}$, the current positions of all vertices $\mathbf{p}_{\text{O}_j}^{\text{v}_k}$ and the current velocity $\mathbf{v}_{\text{O}_j}$ (with $\mathbf{v}_{\text{O}_j}$ being zero for static obstacles). 
All the states of obstacles are observable for all robots.

We define the distributed multi-robot navigation problem as follows: each robot needs to navigate to its goal position within the prescribed time while avoiding collision with other robots and obstacles independently.
We list the notations for this paper in Tab.~\ref{tab:variable_list} for reference.
\begin{table}
    \caption{Notations and Descriptions}
    \label{tab:variable_list}
    \centering
    \resizebox{\linewidth}{!}{
    \begin{tabular}{l|l}
    \hline
    Notation & Descriptions   \\ \hline
    $n$ & Dimension of the space \\ 
    $N$, $M$ & Numbers of robots and obstacles  \\ 
    $\mathbb{N}$, $\mathbb{O}$ & Sets of robots and obstacles \\ 
    $\text{R}_i$, $\text{O}_j$ & $i$th robot and the $j$th obstacle \\ 
    $r_{\text{R}_i}$, $r_{\text{O}_j}$ & Radius of circular-shaped $\text{R}_i$ and $\text{O}_j$ \\ 
    $\mathbf{x}_{\text{R}_i}$, $\mathbf{x}_{\text{O}_j}$ & Current position coordinate of $\text{R}_i$ and $\text{O}_j$  \\
    $\mathbf{x}_{i}^{\text{goal}}$ & Goal position of $\text{R}_i$ \\ 
    $\mathbf{v}_{\text{R}_i}$, $\mathbf{v}_{\text{O}_j}$ & Current velocity of $\text{R}_i$ and $\text{O}_j$  \\
    $\mathbf{p}_{\text{R}_i}^{\text{v}_k}$, $\mathbf{p}_{\text{O}_j}^{\text{v}_h}$ & $k$th, $h$th vertex coordinate of $\text{R}_i$, $\text{O}_j$ \\
    $v_{\text{R}_i, \text{max}}$ & Maximum velocity of $\text{R}_i$ \\
    $\text{CC}_{\text{R}_i|\text{O}_j}$ & Collision cone for $\text{R}_i$ induced by $\text{O}_j$ \\ 
    $\text{VO}_{\text{R}_i|\text{O}_j}(\mathbf{v}_{\text{O}_j})$ & Velocity obstacle for $\text{R}_i$ induced by $\text{O}_j$ \\ 
    $\text{RVO}_{\text{R}_i|\text{R}_j}(\mathbf{v}_{\text{R}_i}, \mathbf{v}_{\text{R}_j})$ & Reciprocal VO for $\text{R}_i$ induced by $\text{R}_j$ \\
    $\mathbf{vl}$, $\mathbf{vr}$ & Direction vectors on both sides of VO \\ 
    $\mathcal{S}_i$, $A_i$, $b_i$ & Set of polytopes  \\
    $\theta_{\mathbf{p}_{\text{O}_j}^{\text{v}_h} - \mathbf{p}_{\text{R}_i}^{\text{v}_k}}$ & Angle between vector $\mathbf{p}_{\text{O}_j}^{\text{v}_h} - \mathbf{p}_{\text{R}_i}^{\text{v}_k}$ and x axis \\
    $\mathbf{e}_x$, $\mathbf{e}_y$ & Unit vectors of x and y axis \\ \hline
    $\Delta t$ & Time step of simulation \\ 
    $t_{\text{max}}$ & Maximum loop time for the simulation \\
    $\mathbf{v}_{\text{R}_i}^{\text{pref}}$, $\mathbf{v}_{\text{R}_i}^{\text{new}}$ & Preferred or new velocity of $\text{R}_i$ \\
    $l$ & Neighboring region of the robot  \\
    $\mathbb{B}_i$, $\mathbb{C}_i$ & Neighboring robots and obstacles of the robot $\text{R}_i$ \\
    ${\text{VO}\,}_{i}$ & Combined velocity obstacle of $\text{R}_i$ \\ 
    ${\text{RVO}\,}_{i}$ & Combined reciprocal velocity obstacle of $\text{R}_i$ \\ 
    $J(\mathbf{v}_{\text{R}_i})$ & Penalty cost function for velocity $\mathbf{v}_{\text{R}_i}$ \\
    $\phi_i$ & Weight coefficient in the penalty function of $\text{R}_i$ \\
    $tc_{i}(\mathbf{v}_{\text{R}_i})$ & Expected collision time when $\text{R}_i$ in velocity $\mathbf{v}_{\text{R}_i}$ \\
    $v$, $w$ & Transitional and rotational velocities \\
    $\text{VO}_{\text{c}}$, $\text{VO}_{\text{p}}$ & Circular-shaped or polytopic-shaped based VO \\ \hline
    \end{tabular}
    }
\end{table}

\subsection{Velocity Obstacle for Circular Shapes}
\label{sec:vo_define}
\begin{figure} 
    \centering
    \includegraphics[width=0.75\linewidth]{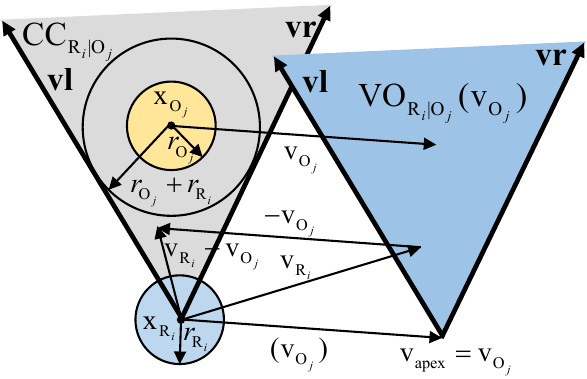}
    \caption{Circular-shaped based velocity obstacle $\text{VO}_{\text{R}_i|\text{O}_j}(\mathbf{v}_{\text{O}_j})$ (blue conic region) of robot $\text{R}_i$ (blue circular region) induced by the obstacle $\text{O}_j$ (yellow circular region) with velocity $\mathbf{v}_{\text{O}_j}$.
    $\text{CC}_{\text{R}_i|\text{O}_j}$ (gray conic region) represents the collision cone between $\text{R}_i$ and $\text{O}_j$.
    If the relative velocity $\mathbf{v}_{\text{R}_i}-\mathbf{v}_{\text{O}_j} \in \text{CC}_{\text{R}_i|\text{O}_j}$ or the absolute robot velocity $\mathbf{v}_{\text{R}_i} \in \text{VO}_{\text{R}_i|\text{O}_j}(\mathbf{v}_{\text{O}_j})$, a collision will occur between $\text{R}_i$ and $\text{O}_j$.
    The direction vectors $\mathbf{vl}$ and $\mathbf{vr}$ (bold solid lines) is same for $\text{CC}_{\text{R}_i|\text{O}_j}$ and $\text{VO}_{\text{R}_i|\text{O}_j}(\mathbf{v}_{\text{O}_j})$ of robot $\text{R}_i$.}
    \hspace{-1.0cm}
    \label{fig:vel_circle}
\end{figure}

% introduce vo
In this section, we present some preliminaries of velocity obstacle representation~\cite{fiorini1998motion}.
For a circular-shaped robot $\text{R}_i$ and a circular-shaped obstacle $\text{O}_j$ with radius $r_{\text{R}_i}$ and $r_{\text{O}_j}$, the current position and velocity could be denoted as $\mathbf{x}_{\text{R}_i}$, $\mathbf{x}_{\text{O}_j}$, $\mathbf{v}_{\text{R}_i}$, $\mathbf{v}_{\text{O}_j}$ respectively. 
Before introducing the concept of VO, we first introduce the concept of collision cone (CC).
For a robot $\text{R}_i$ (with any shape) with velocity $\mathbf{v}_{\text{R}_i}$ and an obstacle $\text{O}_j$ with velocity $\mathbf{v}_{\text{O}_j}$, $\text{CC}_{\text{R}_i|\text{O}_j}$ is defined as follows\cite{fiorini1998motion}:
\begin{equation*}
    \text{CC}_{\text{R}_i|\text{O}_j} = \big\{\mathbf{v}_{\text{R}_i}-\mathbf{v}_{\text{O}_j} \big| \lambda(\mathbf{x}_{\text{R}_i}, \mathbf{v}_{\text{R}_i}-\mathbf{v}_{\text{O}_j}) \cap \text{O}_j \oplus -\text{R}_i \neq \emptyset 
    \big\}
\label{eq:cc_define}
\end{equation*}
where $\lambda(\mathbf{x}, \mathbf{v}) = \{ \mathbf{x} + t \mathbf{v} \big| t>0 \}$ denotes a ray starting at point $\mathbf{x}$ and in the direction of vector $\mathbf{v}$, and $\oplus$ denotes the Minkowski sum, $\text{O}_j \oplus -\text{R}_i$ means considering $\text{R}_i$ as a point with a certain expansion for $\text{O}_j$.
Thus, CC is the set of all relative velocities which lead to a collision.
In particular, if the relative velocity of $\text{R}_i$ and $\text{O}_j$ lies in $\text{CC}_{\text{R}_i|\text{O}_j}$, and both $\text{R}_i$ and $\text{O}_j$ maintain the current velocity, a collision will occur between $\text{R}_i$ and $\text{O}_j$ in the future.
Therefore, any relative velocity outside $\text{CC}_{\text{R}_i|\text{O}_j}$ is guaranteed to be collision free, provided both $\text{R}_i$ and $\text{O}_j$ maintain the current velocity.
The term $\text{CC}_{\text{R}_i|\text{O}_j}$ is defined in terms of relative velocity.
When considering multiple obstacles, it is necessary to establish an equivalent description that uses the absolute velocity of $\text{R}_i$.
Adding the relative velocity $\mathbf{v}_{\text{R}_i}-\mathbf{v}_{\text{O}_j}$ to $\mathbf{v}_{\text{O}_j}$, we can define the convex region $\text{VO}_{\text{R}_i|\text{O}_j}(\mathbf{v}_{\text{O}_j}) = \text{CC}_{\text{R}_i|\text{O}_j} \oplus \mathbf{v}_{\text{O}_j}$, that is
\begin{equation}
\resizebox{0.90\columnwidth}{!}{$
    \text{VO}_{\text{R}_i|\text{O}_j}(\mathbf{v}_{\text{O}_j}) = \big\{
    \mathbf{v}_{\text{R}_i} \big| \lambda(\mathbf{x}_{\text{R}_i}, \mathbf{v}_{\text{R}_i}-\mathbf{v}_{\text{O}_j}) \cap \text{O}_j \oplus -\text{R}_i \neq \emptyset 
    \big\}
$}
\label{eq:vo_define}
\end{equation}
as shown in Fig.~\ref{fig:vel_circle}.
If $\text{R}_i$ holds a constant velocity $\mathbf{v}_{\text{R}_i} \in \text{VO}_{\text{R}_i|\text{O}_j}(\mathbf{v}_{\text{O}_j})$, a collision will occur between $\text{R}_i$ and $\text{O}_j$ in the future, and vice versa.

However, constructing the VO through \eqref{eq:vo_define} is often with high computational complexity because it needs extensive sampling and judgment~\cite{fiorini1998motion}.
In fact, VO defines a geometric conic region of infeasible velocities for the robot.
We can use $\mathbf{vl}$ and $\mathbf{vr}$ to the left and the right side of VO and $\text{v}_\text{apex}$ to represent the apex of the conic region in order to define VO~\cite{han2022reinforcement}, as shown in Fig.~\ref{fig:vel_circle}.
% For collision cone
Moreover, since $\text{VO}_{\text{R}_i|\text{O}_j}(\mathbf{v}_{\text{O}_j})$ is translated from $\text{CC}_{\text{R}_i|\text{O}_j}$, the $\mathbf{vl}$ and $\mathbf{vr}$ of $\text{CC}_{\text{R}_i|\text{O}_j}$ is same as for $\text{VO}_{\text{R}_i|\text{O}_j}(\mathbf{v}_{\text{O}_j})$ and only the apex is different.
The apex $\text{v}_\text{apex}$ of $\text{VO}_{\text{R}_i|\text{O}_j}(\mathbf{v}_{\text{O}_j})$ is at $\mathbf{v}_{\text{O}_j}$, as shown in Fig.~\ref{fig:vel_circle}.
So if we could calculate the direction vectors $\mathbf{vl}$ and $\mathbf{vr}$ on both sides of $\text{CC}_{\text{R}_i|\text{O}_j}$, we can construct $\text{VO}_{\text{R}_i|\text{O}_j}(\mathbf{v}_{\text{O}_j})$ easily.
Most of VO-based works~\cite{van2008reciprocal, berg2011reciprocal, snape2011hybrid} usually choose circular-shaped robots and obstacles, then the direction vectors $\mathbf{vl}$ and $\mathbf{vr}$ on both sides of $\text{CC}_{\text{R}_i|\text{O}_j}$ can be conveniently obtained by calculating the vectors which are tangent to $D(\mathbf{x}_{\text{O}_j}-\mathbf{x}_{\text{R}_i}, r_{\text{R}_i} + r_{\text{O}_j})$, where $D(\mathbf{x}, r)$ is an circle with radius $r$ centered at $\mathbf{x}$.
Then we can construct $\text{VO}_{\text{R}_i|\text{O}_j}(\mathbf{v}_{\text{O}_j})$ with the direction vectors $\mathbf{vl}$ and $\mathbf{vr}$ and the apex.
In this case, the definition~\eqref{eq:vo_define} of VO could be simplified to 
\begin{equation*}
\resizebox{\columnwidth}{!}{$
    \text{VO}_{\text{R}_i|\text{O}_j}(\mathbf{v}_{\text{O}_j}){=}\big\{
    \mathbf{v}_{\text{R}_i} \big| \exists t>0, \, (\mathbf{v}_{\text{R}_i}{-}\mathbf{v}_{\text{O}_j})t \in D(\mathbf{x}_{\text{O}_j}{-}\mathbf{x}_{\text{R}_i}, r_{\text{R}_i}{+}r_{\text{O}_j}) 
    \big\}.
$}
\label{eq:vo_simply}
\end{equation*}
However, the above approach is no longer valid when applied to polytopic-shaped robots, so in this paper we propose constructing the VO for polytopic-shaped robots through an alternative numerically efficient way instead of~\eqref{eq:vo_define}.

\begin{remark}
In fact, $\mathbf{vl}$ and $\mathbf{vr}$ represent the left turn and right turn boundary of the relative velocity between $\text{R}_i$ and $\text{O}_j$. In other words, when two robots are approaching each other, they need to either turn left or right sufficiently to avoid each other.
\end{remark}
\subsection{VO Variants}
\label{sec:vo_variants}
Since VO disregards the reactive nature~\cite{van2008reciprocal} which enables each robot to independently adapt its velocity to avoid collision with other robots and obstacles, this will lead to unnecessary oscillations when using the VO-based algorithm for distributed multi-robot navigation.
Interested readers should refer to ~\cite{van2008reciprocal} for more details.
Some variants of VO explicitly consider this problem and make specific improvements to deal with the unnecessary oscillations, such as Reciprocal Velocity Obstacle (RVO)\cite{van2008reciprocal} and Hybrid Reciprocal Velocity Obstacle (HRVO)~\cite{snape2011hybrid}.
For RVO, it is assumed that both robots take half the responsibility for collision avoidance, and $\text{RVO}_{\text{R}_i|\text{R}_j}(\mathbf{v}_{\text{R}_i}, \mathbf{v}_{\text{R}_j})$ is defined as follows~\cite{van2008reciprocal}:
\begin{equation*}
    \text{RVO}_{\text{R}_i|\text{R}_j}(\mathbf{v}_{\text{R}_i}, \mathbf{v}_{\text{R}_j}) = \big\{ \mathbf{v}^{'}_{\text{R}_i} \big| 2\mathbf{v}^{'}_{\text{R}_i} - \mathbf{v}_{\text{R}_i} \in \text{VO}_{\text{R}_i|\text{R}_j}(\mathbf{v}_{\text{R}_j}) \big\}
\end{equation*}
where $\mathbf{v}_{\text{R}_i}$ and $\mathbf{v}_{\text{R}_j}$ are the current velocity of $\text{R}_i$ and $\text{R}_j$, and $\mathbf{v}^{'}_{\text{R}_i}$ is the new velocity that will lead to a collision.
Each robot chooses a new velocity outside each other's RVO as well as in the same side of each other's RVO, which guarantees collision avoidance.
Since obstacles don't have the reactive nature like robots, RVO is unsuitable to realize collision avoidance between a robot and an obstacle.
The direction vectors $\mathbf{vl}$ and $\mathbf{vr}$ on both sides of RVO is same as VO, and RVO can geometrically be interpreted as VO translated such that its apex lies at $\frac{\mathbf{v}_{\text{R}_i} + \mathbf{v}_{\text{R}_j}}{2}$.
HRVO is proposed to solve the oscillations known as “reciprocal dances" when using RVO for distributed multi-robot navigation, for more details refer to~\cite{snape2011hybrid}.
In summary, the difference between VO, RVO and HRVO is the apex with the direction vectors of these three cones being the same.

\subsection{Obstacle Avoidance Between Polytopes}
\label{sec:obstacle_avoidance_polytope}
For the distributed multi-robot navigation with polytopic shapes, it is necessary to evaluate whether collision exists between robots and obstacles, and this can be described as a minimum distance
problem between polytopes.
Consider two polytopic sets $\mathcal{S}_1$ and $\mathcal{S}_2$, and the distance between these two sets is given by the following primal problem:
\begin{equation}
dist\big( \mathcal{S}_1, \mathcal{S}_2 \big) := \min_{y_1, y_2}  \big\{
\big\| y_1 - y_2 \big\|_2 \big| A_{1} y_1 \leqslant b_1, A_{2} y_2 \leqslant b_2 
\big\}
\label{eq:distance}
\end{equation}
where $\mathcal{S}_1= \big\{ y_1 \in \mathbb{R}^n \big| A_{1} y_1 \leqslant b_1 \big\}$ and $\mathcal{S}_2=\big\{ y_2 \in \mathbb{R}^n \big| A_{2} y_2 \leqslant b_2 \big\}$, $A_1$, $b_1$, $A_2$ and $b_2$ depend on the robot's and obstacle's positions.
$n$ represents the space dimension and is considered as $n = 2$ in the rest of this paper.
Thus, when the polytopes don't collide with each other, the minimum distance between two polytopes is positive, i.e., $dist\big(\mathcal{S}_1, \mathcal{S}_2\big) > 0$.
So we can use \eqref{eq:distance} to check if there is a collision between robots and obstacles with polytopic-shaped.
\section{Multi-Robot Navigation with Polytopic-shaped Robots}
\label{sec:multi}
\begin{figure}
    \centering
    \includegraphics[width=0.80\linewidth]{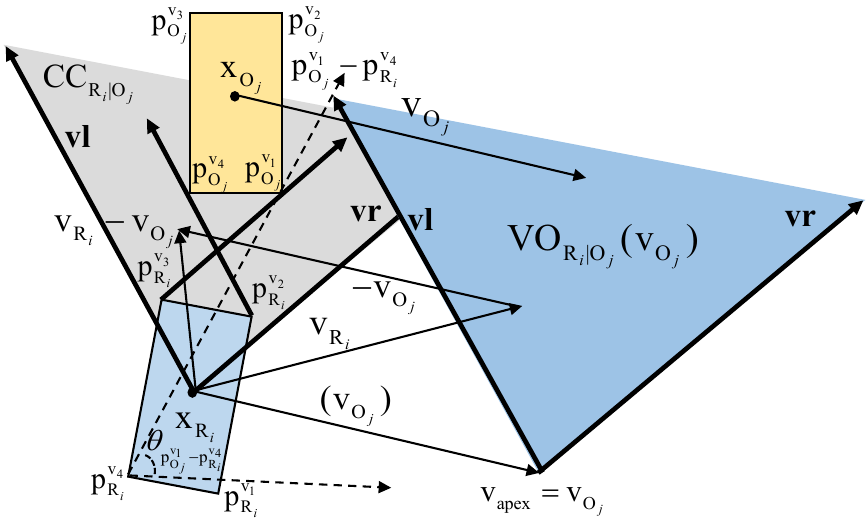}
    \caption{Velocity Obstacle $\text{VO}_{\text{R}_i|\text{O}_j}(\mathbf{v}_{\text{O}_j})$ of robot $\text{R}_i$ induced by the obstacle $\text{O}_j$ for polytopic-shaped robots. 
    First, we need to obtain the direction vectors $\mathbf{vl}$ and $\mathbf{vr}$ (bold black solid lines) on both sides of the collision cone $\text{CC}_{\text{R}_i|\text{O}_j}$ by connecting the vertices of $\text{R}_i$ and $\text{O}_j$ in any two pairs as done in~\eqref{eq:vl_and_vr}.
    Then we construct the $\text{VO}_{\text{R}_i|\text{O}_j}(\mathbf{v}_{\text{O}_j})$ with the direction vectors $\mathbf{vl}$ and $\mathbf{vr}$, and $\text{VO}_{\text{R}_i|\text{O}_j}(\mathbf{v}_{\text{O}_j})$ is a cone with its apex at $\mathbf{v}_{\text{O}_j}$.}
    \label{fig:vel_polytope}
\end{figure}
In this section, we firstly propose an optimization-free approach to construct the VO for polytopic-shaped robots in Sec.~\ref{sec:polytope_vo}.
Then, we demonstrate how to realize the distributed multi-robot navigation with polytopic shapes using VO and its variants in Sec.~\ref{sec:navigation_alg}.
\subsection{Velocity Obstacle for Polytopic-shaped Robots}
\label{sec:polytope_vo}
% introduce the VO for polytopic robot
In Sec.~\ref{sec:vo_define}, we have reviewed the concept of VO and demonstrated how to construct the VO for circular-shaped robots. 
In the following, we will extend the concept of VO to polytopic-shaped robots and obstacles and demonstrate how to construct the VO for polytopic-shaped robots with high computational efficiency.

% method for construct VO for polytopes
In order to construct VO for polytopic-shaped robots, we are going to use the vertex coordinates, together with the robot's other external states, including robot's current position and velocity.
Assume there is a robot $\text{R}_i$ with $K_{\text{R}_i}$ vertices and an obstacle $\text{O}_j$ with $K_{\text{O}_j}$ vertices in the shared environment, and the current position, current velocity, and each vertex's current coordinate of $\text{R}_i$ and $\text{O}_j$ could be denoted as $\mathbf{x}_{\text{R}_i}$, $\mathbf{x}_{\text{O}_j}$, $\mathbf{v}_{\text{R}_i}$, $\mathbf{v}_{\text{O}_j}$, $\mathbf{p}_{\text{R}_i}^{\text{v}_k}$, $\mathbf{p}_{\text{O}_j}^{\text{v}_h}$, respectively.
As we mentioned in Sec.~\ref{sec:vo_define}, calculating the direction vectors $\mathbf{vl}$ and $\mathbf{vr}$ on both side of the cone $\text{VO}_{\text{R}_i|\text{O}_j}(\mathbf{v}_{\text{O}_j})$ is necessary and sufficient to construct $\text{VO}_{\text{R}_i|\text{O}_j}(\mathbf{v}_{\text{O}_j})$, and in the following we will demonstrate how to calculate it.

For two polytopes $\text{R}_i$ and $\text{O}_j$, a vertex pair $(\mathbf{p}_{\text{O}_j}^{\text{v}_h}, \mathbf{p}_{\text{R}_i}^{\text{v}_k})$ could be obtained by selecting any two vertices of $\text{O}_j$ and $\text{R}_i$, where $h \in \{1, 2, \dots, K_{\text{O}_j}\}$ and $k \in \{1, 2, \dots, K_{\text{R}_i}\}$, and there are $K_{\text{R}_i} \times K_{\text{O}_j}$ vertex pairs in total.
By connecting the two vertices in a vertex pair $(\mathbf{p}_{\text{O}_j}^{\text{v}_h}, \mathbf{p}_{\text{R}_i}^{\text{v}_k})$ as a vector, we could obtain the angle between the vector $\mathbf{p}_{\text{O}_j}^{\text{v}_h} - \mathbf{p}_{\text{R}_i}^{\text{v}_k}$ and the x axis:
% {\small
\begin{equation*}
    \theta_{\mathbf{p}_{\text{O}_j}^{\text{v}_h} - \mathbf{p}_{\text{R}_i}^{\text{v}_k}} = \tan^{-1}[(\mathbf{p}_{\text{O}_j}^{\text{v}_h} - \mathbf{p}_{\text{R}_i}^{\text{v}_k}) \cdot \mathbf{e}_y / (\mathbf{p}_{\text{O}_j}^{\text{v}_h} - \mathbf{p}_{\text{R}_i}^{\text{v}_k}) \cdot \mathbf{e}_x ] \in [-\pi, \pi]
    \label{eq:theta_calculate_scope}
\end{equation*}
% }
as shown in Fig.~\ref{fig:vel_polytope}.
Then we could obtain the direction vectors $\mathbf{vl}$ and $\mathbf{vr}$ on both sides of the cone as follows,
\begin{equation}
    \begin{aligned}
    \mathbf{vl} = \cos(\max \limits_{\text{v}_h, \text{v}_k} \: \theta_{\mathbf{p}_{\text{O}_j}^{\text{v}_h} - \mathbf{p}_{\text{R}_i}^{\text{v}_k}}) \mathbf{e}_x + \sin(\max \limits_{\text{v}_h, \text{v}_k} \: \theta_{\mathbf{p}_{\text{O}_j}^{\text{v}_h} - \mathbf{p}_{\text{R}_i}^{\text{v}_k}}) \mathbf{e}_y \\ 
    \mathbf{vr} = \cos(\min \limits_{\text{v}_h, \text{v}_k} \: \theta_{\mathbf{p}_{\text{O}_j}^{\text{v}_h} - \mathbf{p}_{\text{R}_i}^{\text{v}_k}}) \mathbf{e}_x + \sin(\min \limits_{\text{v}_h, \text{v}_k} \: \theta_{\mathbf{p}_{\text{O}_j}^{\text{v}_h} - \mathbf{p}_{\text{R}_i}^{\text{v}_k}}) \mathbf{e}_y \\ 
    \end{aligned}
    \label{eq:vl_and_vr}
\end{equation}
where $\mathbf{vl}$ represents the unit vector corresponding to the largest angle with respect to the x axis among all the vertex pairs, and $\mathbf{vr}$ represents the smallest one, as shown in Fig.~\ref{fig:vel_polytope}.

After calculating the unit vectors $\mathbf{vl}$ and $\mathbf{vr}$ on both sides of the cone, and as discussed in Sec.~\ref{sec:vo_define}, $\text{VO}_{\text{R}_i|\text{O}_j}(\mathbf{v}_{\text{O}_j})$ can be then easily constructed as a cone with its apex at $\mathbf{v}_{\text{O}_j}$, shown in Fig.~\ref{fig:vel_polytope}.

\subsection{Navigation of Distributed Multi-Robot System}
\label{sec:navigation_alg}
% how to use VO to realize the navigation
This section mainly demonstrates how to achieve the distributed multi-robot navigation with polytopic shapes and the definition of this problem can be found in Sec.~\ref{sec: problem statement}.

\begin{algorithm}
\caption{Distributed Navigation for Multi-Robot Systems with VO}
\begin{algorithmic}
\label{alg:multi_robot}
    \renewcommand{\algorithmicrequire}{\textbf{Initialization:}}
    \REQUIRE All robots' start positions and goal positions.
    \STATE
    Initialize all robots states and set $t=0$.
    \WHILE{$t \leq t_{\text{max}}$ \AND at least one robot doesn't arrive at the goal position and is not stopped}
        \FOR{$\text{R}_i \in \mathbb{N}$}
            \STATE
            Calculate the new velocity $\mathbf{v}_{\text{R}_i}^{\text{new}}$ for each robot $\text{R}_i$ according \eqref{eq:v_choose} and \eqref{eq:v_choose_crowded}. 
        \ENDFOR
        \FOR{$\text{R}_i \in \mathbb{N}$}
            \STATE
            Update the position $\mathbf{x}_{\text{R}_i}$ with robot dynamics.
        \ENDFOR
        \FOR{$\text{R}_i \in \mathbb{N}$}
            \STATE
            Check the minimum distance with other robots and obstacles \eqref{eq:distance}, and stop robot $\text{R}_i$ if minimum distance is zero.
        \ENDFOR
        \STATE
        $t = t + \Delta t$.
    \ENDWHILE
\end{algorithmic}
\end{algorithm}

For the distributed multi-robot navigation with polytopic shapes, we assume that all robots navigate to their goal positions using the same policy, and the pseudocode of the overall method is shown in Alg.~\ref{alg:multi_robot}.
The most critical part of the whole algorithm is about how to select a new velocity $\mathbf{v}_{\text{R}_i}^{\text{new}}$ for each robot $\text{R}_i$.
In the following we will demonstrate how to select the new velocity $\mathbf{v}_{\text{R}_i}^{\text{new}}$.

\subsubsection{Construct Combined Velocity Obstacle for Each Robot}
% introduce how to construct the combined VO for each robot
We have introduced $\text{VO}_{\text{R}_i|\text{O}_j}(\mathbf{v}_{\text{O}_j})$ to achieve collision avoidance between the robot $\text{R}_i$ and the obstacle $\text{O}_j$ by choosing a velocity $\mathbf{v}_{\text{R}_i}$ which is outside of $\text{VO}_{\text{R}_i|\text{O}_j}(\mathbf{v}_{\text{O}_j})$.
During the navigation progress, if a robot $\text{R}_i$ needs to avoid collision with all other robots and obstacles around it, it is necessary to select a velocity outside of all the VO induced by each robot or obstacle.
Before introducing the combined velocity obstacle~\cite{van2008reciprocal}, we first define the set of neighboring robots and obstacles around the robot $\text{R}_i$ as $\mathbb{B}_i$ and $\mathbb{C}_i$, respectively.
% For instance, if we set $\mathbb{B}_i=\{j|\text{R}_j \in \mathbb{N}, j  \neq i\}$ and $\mathbb{C}_i=\{j|\text{O}_j  \in \mathbb{O} \}$, which means all the robots (expect itself) as well as obstacles are considered.
% To reduce the computational burden in practical applications, only 
The neighboring robots and obstacles around the current position of $\text{R}_i$ within a distance of magnitude $l$ will be taken into account, and we set $\mathbb{B}_i=\{j|\text{R}_j \in \mathbb{N}, j  \neq i, \| \mathbf{x}_{\text{R}_i} - \mathbf{x}_{\text{R}_j} \|_2 \leq l \}$ and $\mathbb{C}_i=\{j|\text{O}_j \in \mathbb{O}, \| \mathbf{x}_{\text{R}_i} - \mathbf{x}_{\text{O}_j} \|_2 \leq l \}$.
$l$ is a variable representing the neighboring region, as shown in Fig.~\ref{fig:environment}, and the optimal $l$ depends on the maximum velocity of each robot and the obstacle as well as the time $\tau$, which represents the time window for collision-free motion.
For any robot $\text{R}_i$ with the obstacle $\text{O}_j$, the optimal neighboring region could be set as~\eqref{eq:neighboring_region}:
\begin{equation}
    l=(v_{\text{R}_i, \text{max}} + v_{\text{O}_j, \text{max}}) \tau
    \label{eq:neighboring_region}
\end{equation}

Then we define the combined velocity obstacle for robot $\text{R}_i$ as~\eqref{eq:com_vo}:
\begin{equation}
    {\text{VO}}_{i}=\bigcup_{j \in \mathbb{B}_i} \: \text{VO}_{\text{R}_i \, | \, \text{R}_j} \: \cup \: \bigcup_{j \in \mathbb{C}_i} \: \text{VO}_{\text{R}_i \, | \,\text{O}_j}
    \label{eq:com_vo}
\end{equation}
which is a union of all the VO induced by each robot or obstacle around the robot $\text{R}_i$. 
Therefore, each robot $\text{R}_i$ could realize collision avoidance with other robots and obstacles by choosing a velocity outside of its combined velocity obstacle $\text{VO}_i$.

In order to implement RVO and HRVO for polytopic-shaped robots, the combined reciprocal velocity obstacle and the combined hybrid reciprocal velocity obstacle are defined as follows:
\begin{equation}
\begin{gathered}
    {\text{RVO}}_{i}=\bigcup_{j \in \mathbb{B}_i} \: \text{RVO}_{\text{R}_i \, | \, \text{R}_j} \: \cup \: \bigcup_{j \in \mathbb{C}_i} \: \text{VO}_{\text{R}_i \, | \,\text{O}_j} \\
    {\text{HRVO}}_{i}=\bigcup_{j \in \mathbb{B}_i} \: \text{HRVO}_{\text{R}_i \, | \, \text{R}_j} \: \cup \: \bigcup_{j \in \mathbb{C}_i} \: \text{VO}_{\text{R}_i \, | \,\text{O}_j}
\end{gathered}
\end{equation}
where VO induced by the other robots are replaced by RVO or HRVO to reduce unnecessary oscillation.

\subsubsection{Selecting Velocity for Each Robot}
\begin{figure} 
    \centering
    \includegraphics[width=0.9\linewidth]{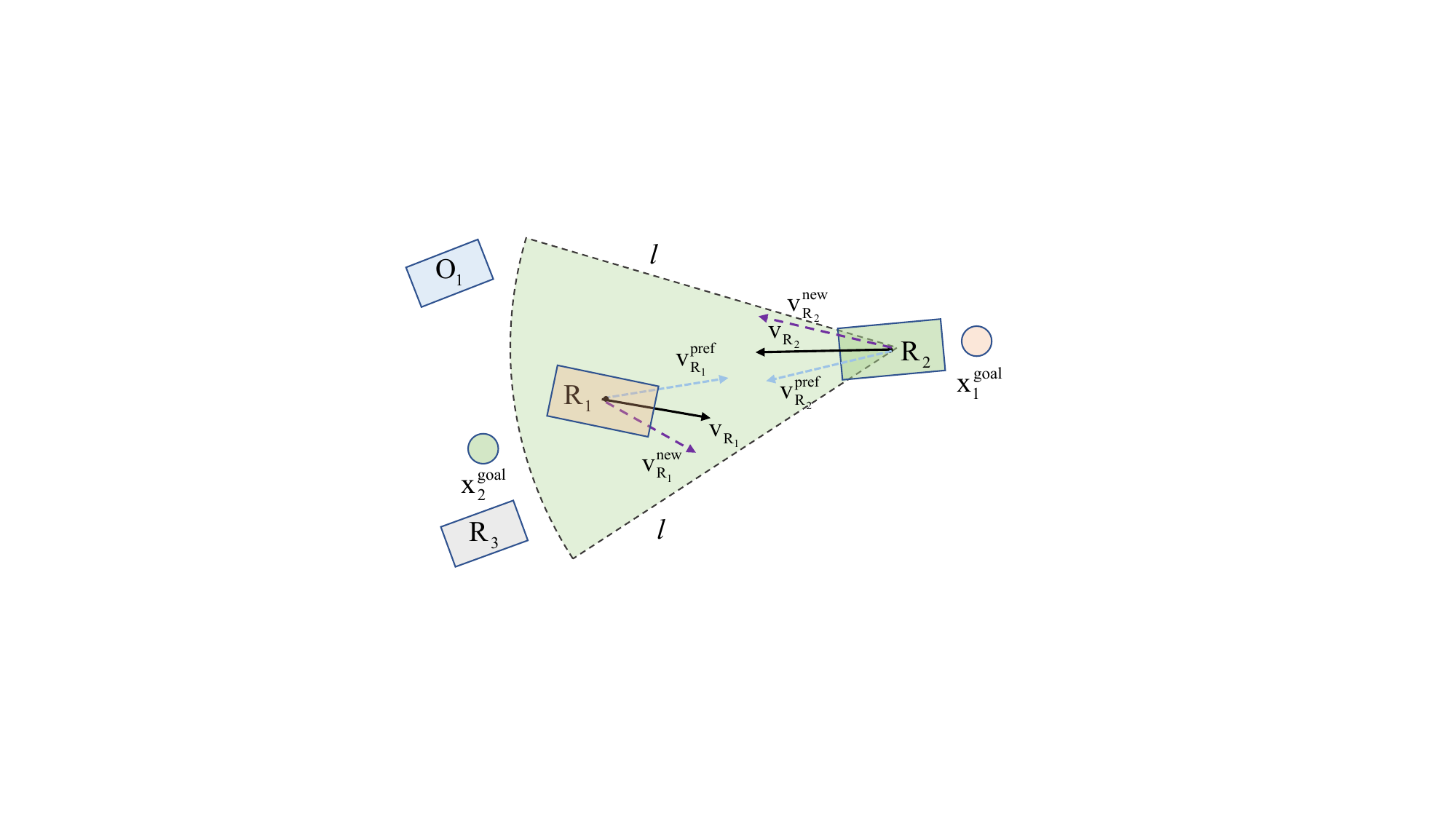}
    \caption{Illustration for the distributed navigation for multi-robot system. Robot $\text{R}_2$ only considers the robots and obstacles within a certain range ($l$), so $\text{R}_2$ only considers the VO induced by $\text{R}_1$.}
    \label{fig:environment}
\end{figure}
To realize the distributed multi-robot navigation with polytopic shapes, the selected velocity not only needs to avoid collision with other robots and obstacles but also needs to guide the robot to its goal position.
So we could choose a new velocity $\mathbf{v}_{\text{R}_i}^{\text{new}}$ that is outside of the combined velocity obstacle $\text{VO}_i$ and as close as possible to the preferred velocity for each robot $\text{R}_i$ independently as follows,
\begin{equation}
    \begin{split}
    & \mathbf{v}_{\text{R}_i}^{\text{new}} = \argmin_{\mathbf{v}_{\text{R}_i} \notin \text{VO}_i} \, 
    \| \mathbf{v}_{\text{R}_i} - \mathbf{v}_{\text{R}_i}^{\text{pref}} \|_2, \\
    & \mathbf{v}_{\text{R}_i}^{\text{pref}} = v_{\text{R}_i, \text{max}} \frac{\mathbf{x}_{\text{R}_i} - \mathbf{x}_{i}^{\text{goal}}}{\| \mathbf{x}_{\text{R}_i} - \mathbf{x}_{i}^{\text{goal}} \|_2},
    \end{split}
\label{eq:v_choose}
\end{equation}
where the direction of the preferred velocity $\mathbf{v}_{\text{R}_i}^{\text{pref}}$ is from the current position $\mathbf{x}_{\text{R}_i}$ to the goal position $\mathbf{x}_{i}^{\text{goal}}$ to guide the robot to its goal position as fast as possible, shown in Fig.~\ref{fig:environment}.
\begin{remark}
\label{rem:global-planner}
When the environment is crowded with many static obstacles, selecting a preferred velocity as above may result in deadlock, which could be improved by a global path planner, such as RRT*\cite{karaman2010optimal}. Notice that the global path planner configuration is out of scope of this paper.
\end{remark}
However, there may be no feasible solution for~\eqref{eq:v_choose} in extremely crowded environment, where the combined velocity obstacle $\text{VO}_i$ saturates the entire velocity space and picking a velocity outside $\text{VO}_i$ is impossible.
Moreover, choosing any velocity in this case will result in a collision, only the expected collision time is different.
Therefore we need to make a trade-off between safety maneuver and traveling to the goal position as quickly as possible.
We can select the new velocity $\mathbf{v}_{\text{R}_i}^{\text{new}}$ inside the combined velocity obstacle $\text{VO}_i$ as follows,
\begin{equation}
    \mathbf{v}_{\text{R}_i}^{\text{new}} = \argmin_{\mathbf{v}_{\text{R}_i} \in \text{VO}_i} \, J(\mathbf{v}_{\text{R}_i}),
    J(\mathbf{v}_{\text{R}_i}) = \phi_i \, \frac{1}{tc_{i}(\mathbf{v}_{\text{R}_i})} +
    \| \mathbf{v}_{\text{R}_i} - \mathbf{v}_{\text{R}_i}^{\text{pref}} \|_2
\label{eq:v_choose_crowded}
\end{equation}
where $J(v_{\text{R}_i})$ is a penalty cost function for velocity, the penalty of the candidate velocity depends on the expected collision time $tc_{i}(\mathbf{v}_{\text{R}_i})$ and the difference between the candidate velocity and the preferred velocity $\mathbf{v}_{\text{R}_i}^{\text{pref}}$, $\phi_i$ is the weight coefficient between these two components for each robot $\text{R}_i$, and we choose the velocity with the minimum penalty cost as the new velocity $\mathbf{v}_{\text{R}_i}^{\text{new}}$.
For the expected collision time $tc_{i}(\mathbf{v}_{\text{R}_i})$, we calculate it as follows,
\begin{equation}
    tc_{i}(\mathbf{v}_{\text{R}_i}) = \min \limits_{\text{R}_j \in \mathbb{B}_i, \text{O}_j \in \mathbb{C}_i} \: \{tc_{\text{R}_i}^{\text{R}_j}(\mathbf{v}_{\text{R}_i}), tc_{\text{R}_i}^{\text{O}_j}(\mathbf{v}_{\text{R}_i})\}
\end{equation}
where $tc_{\text{R}_i}^{\text{R}_j}$ or $tc_{\text{R}_i}^{\text{O}_j}$ is the expected collision time estimated from the current position and the velocity of the robot $\text{R}_i$ for collision with robot $\text{R}_j$ or obstacle $\text{O}_j$.
If there is no collision, the expected collision time will be set as infinity.

\section{Simulation Results}
\label{sec:exper}
\subsection{Simulation Setup}
We use an Ubuntu Laptop with Intel Core i7-9750H (CPU 4.5 GHz), NVIDIA GTX 1650 (GPU, 1665MHz) and the whole simulation environment is setup with Python for all computations.
The size of the simulation scene is $\si[per-mode=symbol]{10 \, \metre \times 10 \, \metre}$, the time step $\Delta t$ of simulation is set as $0.1 \, \si[per-mode=symbol]{\second}$, and the maximum time $t_{\text{max}}$ of loop is set as $30\, \si[per-mode=symbol]{\second}$.
In our simulation trails, all the robots are non-holonomic and controlled by the transitional speed $v$ and angular speed $\omega$.
However, the new velocity $\mathbf{v}_{\text{R}_i}^{\text{new}}$ selected by the VO-based navigation algorithm is the orthogonal velocity $\mathbf{v}_{\text{R}_i}^{\text{new}} = [ \mathbf{v}_{x}^{\text{new}}, \mathbf{v}_{y}^{\text{new}} ]$.
To convert to a form which can be used by non-holonomic robots, the orthogonal velocity $\mathbf{v}_{\text{R}_i}^{\text{new}}$ is converted to transitional speed $v$ and angular speed $\omega$ as follows:
\begin{equation*}
    v = \| \mathbf{v}_{\text{R}_i}^{\text{new}} \| \cdot \cos{\varsigma}, \quad \omega = -\varsigma / \eta
\end{equation*}
where $\varsigma$ is the orientation difference between the robot's current orientation and the direction of the orthogonal velocity $\mathbf{v}_{\text{R}_i}^{\text{new}}$, and $\eta$ is the time to adjust the orientation difference to $0$.
The simulation parameters associated with the robot are listed in Tab.~\ref{tab:simulation_para}.
\begin{table}
\caption{Setup of the simulation parameter associated with robot}
\label{tab:simulation_para}
\centering
\begin{tabular}{l|l|l}
\hline
Notation & Meaning     & Value     \\ \hline
$v_{\text{R}_i, \text{max}}$ & Robot's maximum orthogonal velocity & $1.5 \, \si[per-mode=symbol]{\metre\per\second}$ \\
$v_{\text{max}}$ & Robot's maximum transitional speed & $1.5 \, \si[per-mode=symbol]{\metre\per\second}$ \\
$w_{\text{max}}$ & Robot's maximum angular speed & $1.0 \, \si[per-mode=symbol]{\radian\per\second}$ \\
$\eta$ & Time of adjusting orientation difference & $0.2 \, \si[per-mode=symbol]{\second}$ \\
$l$ & Robot's neighboring region & $5.0 \, \si[per-mode=symbol]{\metre}$ \\
$\phi$ & Weight coefficient of the penalty function & $4.0$ \\ \hline     \end{tabular}%
\end{table}
Moreover, a safe margin of $0.15 \, \si[per-mode=symbol]{\metre}$ is added to each polytopic robot to ensure tolerance for motion integration errors and to guarantee a minimum safety distance.

To distinguish with the circular-shaped based VO and for the sake of simplicity, in the following part we denote circular-shaped based VO as $\text{VO}_\text{c}$~\cite{fiorini1998motion}, our polytopic-shaped based VO as $\text{VO}_\text{p}$, circular-shaped based RVO as $\text{RVO}_\text{c}$~\cite{van2008reciprocal}, our polytopic-shaped based RVO as $\text{RVO}_\text{p}$, circular-shaped based HRVO as $\text{HRVO}_\text{c}$~\cite{snape2011hybrid}, and our polytopic-shaped based HRVO as $\text{HRVO}_\text{p}$.
In the following, we will validate the performance of our approach for distributed multi-robot navigation with polytopic shapes in many challenging scenarios.

\subsection{Navigation of Distributed Multi-robot System in Circle Scenario}
\begin{figure}
    \centering
    \begin{subfigure}{0.49\linewidth}
        \centering
        \includegraphics[width=0.95\linewidth]{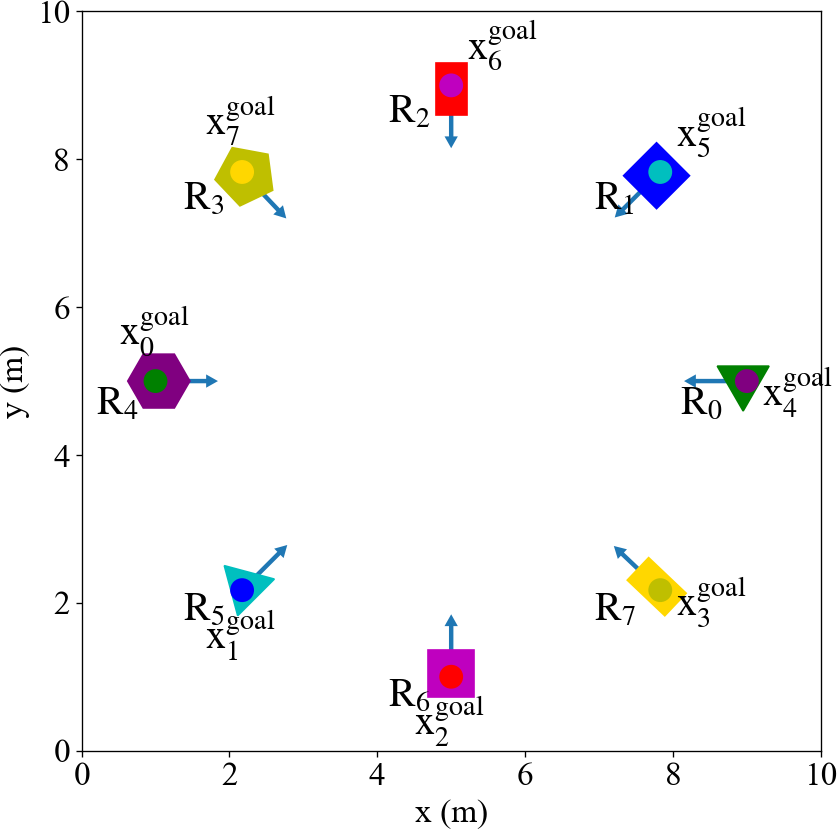}
        \caption{Simulation setup}
        \label{subfig:initial position}
    \end{subfigure}
    \centering
    \begin{subfigure}{0.49\linewidth}
        \centering
        \includegraphics[width=0.95\linewidth]{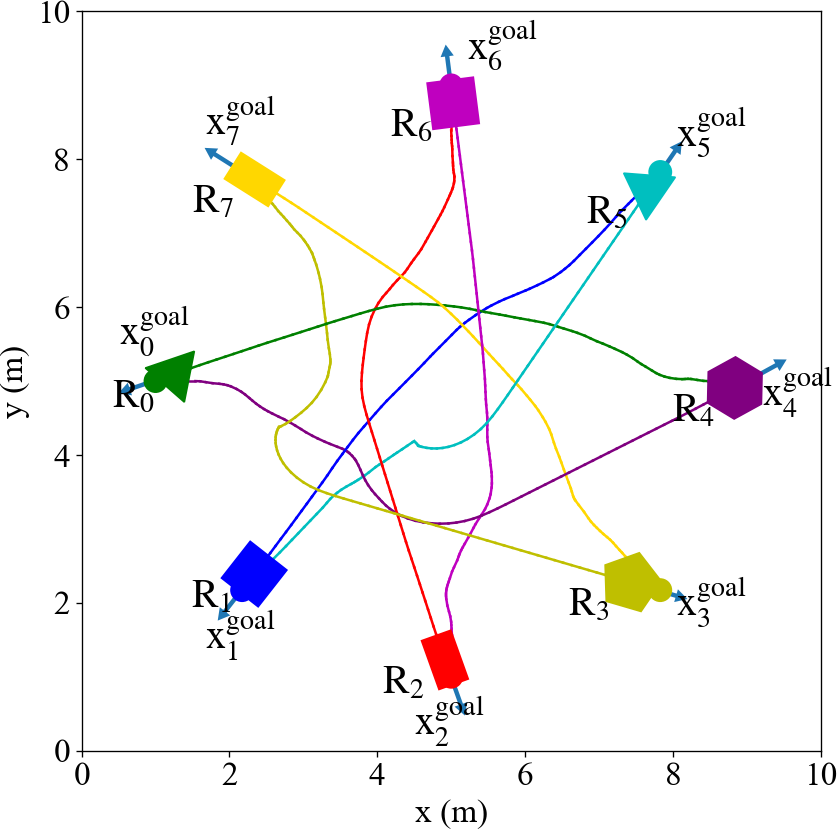}
        \caption{$\text{VO}_\text{p}$}
        \label{subfig:VO_Eight}
    \end{subfigure}
    
    \centering
    \begin{subfigure}{0.49\linewidth}
        \centering
        \includegraphics[width=0.95\linewidth]{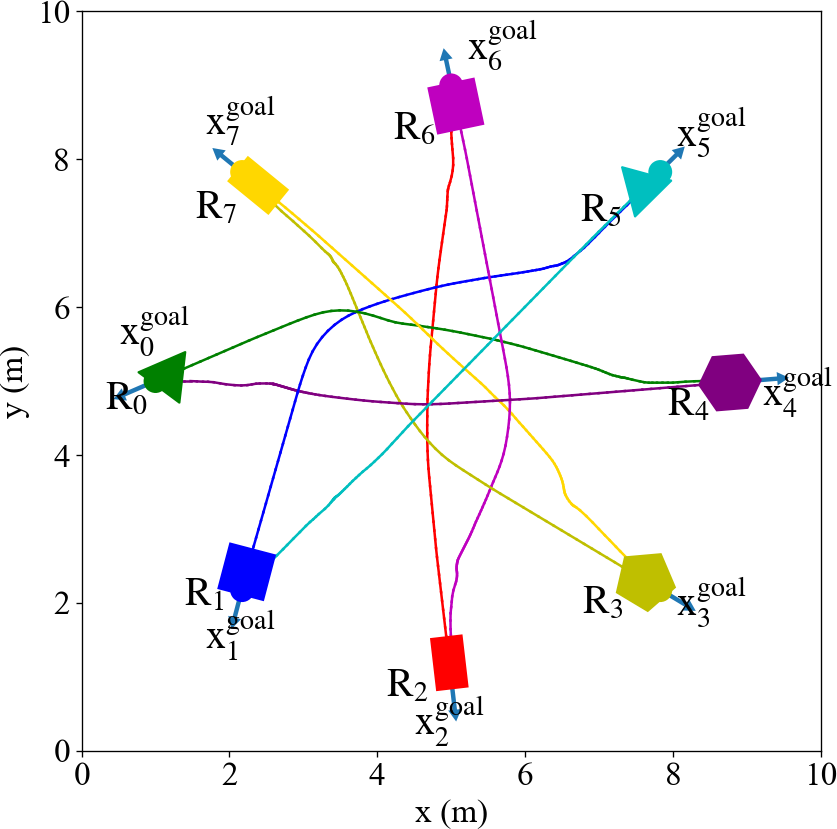}
        \caption{$\text{RVO}_\text{p}$}
        \label{subfig:RVO_Eight}
    \end{subfigure}
    \centering
    \begin{subfigure}{0.49\linewidth}
        \centering
        \includegraphics[width=0.95\linewidth]{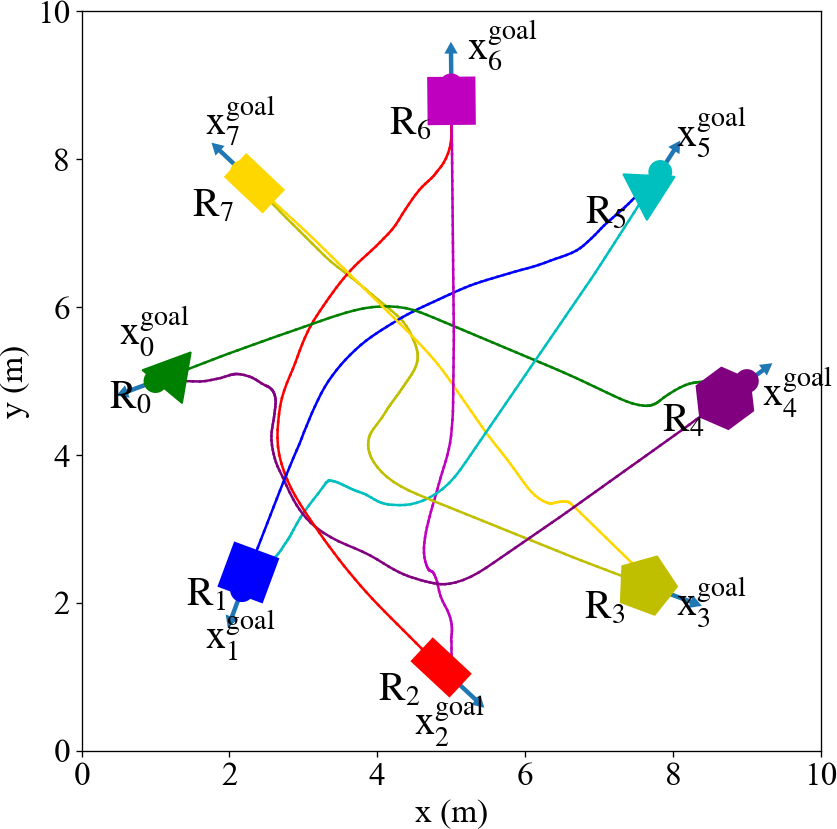}
        \caption{$\text{HRVO}_\text{p}$}
        \label{subfig:HRVO_Eight}
    \end{subfigure}
    \caption{Simulation results of eight polytopic-shaped robots in a circle scenario using our proposed approach under (b) polytopic-shaped based velocity obstacle ($\text{VO}_\text{p}$), (c) polytopic-shaped based reciprocal velocity obstacle ($\text{RVO}_\text{p}$), (d) polytopic-shaped based hybrid reciprocal velocity obstacle ($\text{HRVO}_\text{p}$) representations. In (a), we show the initial and goal position of each robot in a circle scenario, where the goal position of each robot is shown as a circle with different colors.}
    \hspace{-0.5cm}
    \label{fig:eight_robots_simulation}
\end{figure}
We first validate the approach for distributed multi-robot navigation with polytopic shapes in the circle scenario: a certain number of polytopic-shaped robots are uniformly located on a circle of radius $4 \, \si[per-mode=symbol]{\metre}$ centered at $(5 \, \si[per-mode=symbol]{\metre}, 5 \, \si[per-mode=symbol]{\metre})$, and the initial and goal position of robots are symmetric along the center of the circle, as shown in Fig.~\ref{subfig:initial position}.
Our approach for polytopic obstacle avoidance with $\text{VO}_\text{p}$, together with its variants on $\text{RVO}_\text{p}$ and $\text{HRVO}_\text{p}$, are analyzed in the numerical simulation trails. The simulation results of the navigation task for eight heterogeneous polytopic-shaped robots with a circular configuration are shown in Fig.~\ref{fig:eight_robots_simulation}, where all our navigation algorithms move each robot to its goal position while avoiding collision with other robots.
The trajectories generated by $\text{VO}_\text{p}$, $\text{RVO}_\text{p}$ and $\text{HRVO}_\text{p}$ are also illustrated in Fig.~\ref{fig:eight_robots_simulation}, where $\text{RVO}_\text{p}$ and $\text{HRVO}_\text{p}$ have fewer unnecessary oscillations compared with $\text{VO}_\text{p}$.
Moreover, our approach could achieve the distributed multi-robot navigation with polytopic shapes in a large-scaled multi-robot systems with up to $16$ robots, for more details readers can refer to the attached video.

\subsection{Navigation with Static and Dynamic Obstacles}
\begin{figure}
    % \centering
    % \begin{subfigure}{0.49\linewidth}
    %     \centering
    %     \includegraphics[width=0.95\linewidth]{figures/obstacle1.png}
    %     \caption{$t = 0 \, \si[per-mode=symbol]{\second}$}
    %     \label{subfig:obstacle_case1}
    % \end{subfigure}
    \centering
    \begin{subfigure}{0.49\linewidth}
        \centering
        \includegraphics[width=0.95\linewidth]{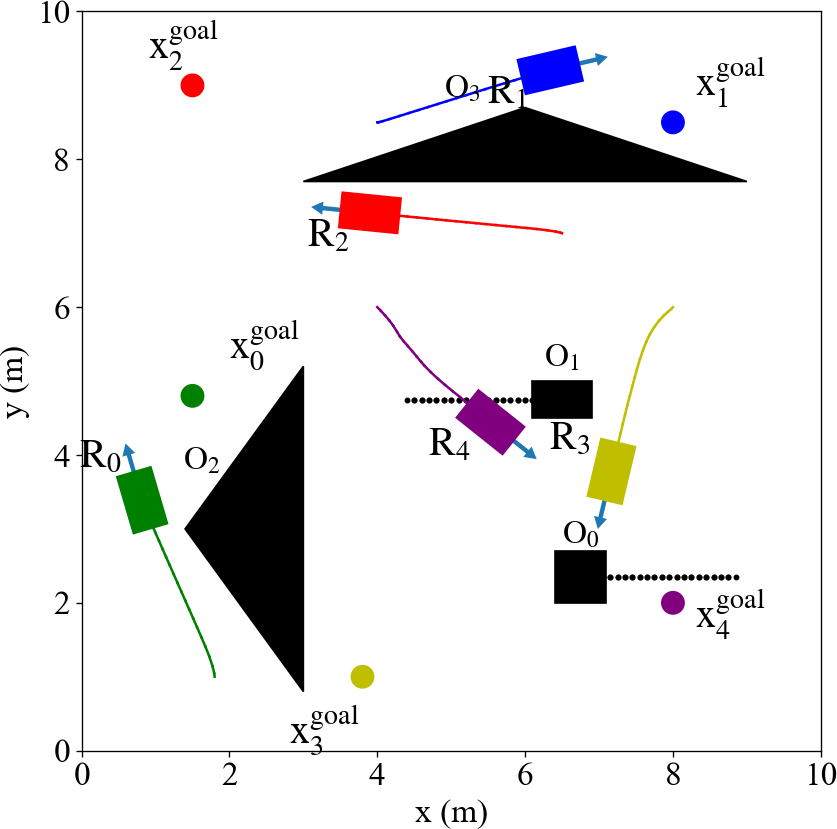}
        \caption{$t = 2.1 \, \si[per-mode=symbol]{\second}$}
        \label{subfig:obstacle_case2}
    \end{subfigure}
    % \centering
    % \begin{subfigure}{0.49\linewidth}
    %     \centering
    %     \includegraphics[width=0.95\linewidth]{figures/obstacle3.png}
    %     \caption{$t = 3.1 \, \si[per-mode=symbol]{\second}$}
    %     \label{subfig:obstacle_case3}
    % \end{subfigure}
    \centering
    \begin{subfigure}{0.49\linewidth}
        \centering
        \includegraphics[width=0.95\linewidth]{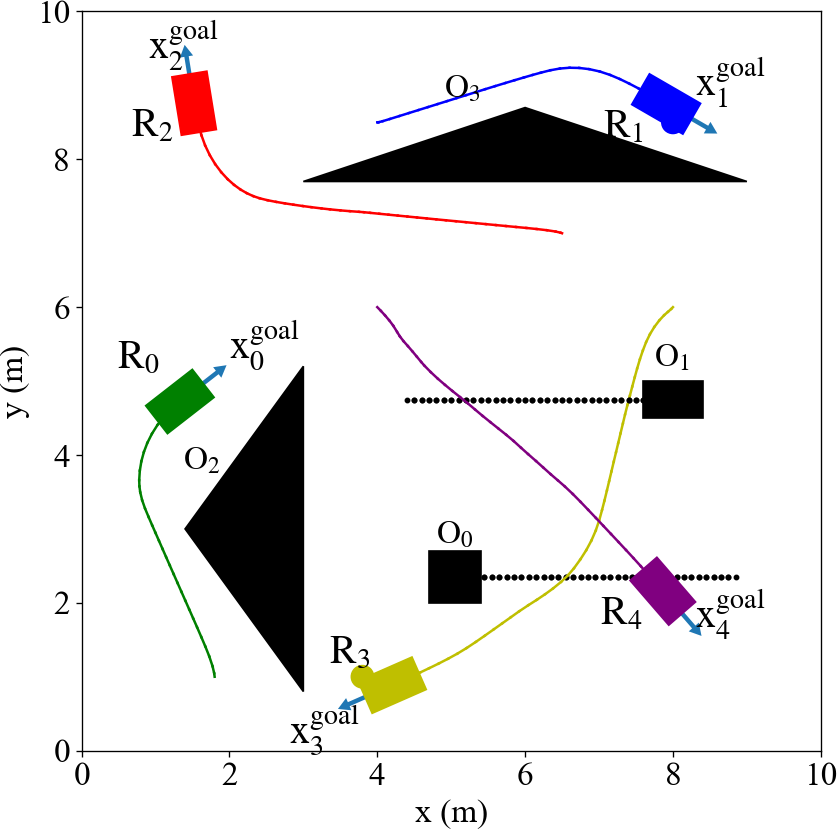}
        \caption{$t = 5.2 \, \si[per-mode=symbol]{\second}$}
        \label{subfig:obstacle_case4}
    \end{subfigure}
    \caption{The navigation simulation results for $5$ polytopic-shaped robots with $2$ static and $2$ dynamic obstacles at two different moments, with the algorithm based on $\text{RVO}_\text{p}$. The black polytopes represent obstacles, and the traveled trajectories of dynamic obstacles are marked as dotted lines.}
    \label{fig:obstacle}
\end{figure}
We also validate the approach for the distributed multi-robot navigation with polytopic shapes in the scenario with static and dynamic polytopic-shaped obstacles.
In this scenario, the initial and goal position of each robot are manually set.
Fig.~\ref{fig:obstacle} shows the navigation simulation results for $5$ polytopic-shaped robots with $2$ static and $2$ dynamic obstacles in the shared environment.
The dynamic obstacles travel at a speed of $1.0 \, \si[per-mode=symbol]{\metre\per\second}$: one dynamic obstacle travels from $(8.85 \, \si[per-mode=symbol]{\metre}, 2.35 \, \si[per-mode=symbol]{\metre})$ to $(5.0 \, \si[per-mode=symbol]{\metre}, 2.35 \, \si[per-mode=symbol]{\metre})$ and the other travels from $(4.4 \, \si[per-mode=symbol]{\metre}, 4.75 \, \si[per-mode=symbol]{\metre})$ to $(8.0 \, \si[per-mode=symbol]{\metre}, 4.75 \, \si[per-mode=symbol]{\metre})$.
According to Fig.~\ref{fig:obstacle}, we observe that each robot successfully navigates to its goal position while avoiding collision with other robots and obstacles using the navigation algorithm based on $\text{RVO}_\text{p}$.
As shown in Fig.~\ref{subfig:obstacle_case2} and Fig.~\ref{subfig:obstacle_case4}, the robots $\text{R}_0$, $\text{R}_1$ and $\text{R}_2$ explicitly adjust their velocities to avoid collisions with the static obstacles, and $\text{R}_3$ and $\text{R}_4$ explicitly adjust their velocity to avoid collisions with the dynamic obstacles and other robots.

\begin{table*}[t]
\caption{A performance comparison of proposed approaches (polytopic-shaped based VO ($\text{VO}_\text{p}$), RVO ($\text{RVO}_\text{p}$), HRVO ($\text{HRVO}_\text{p}$)) with the state of the art (circular-shaped based VO ($\text{VO}_\text{c}$)~\cite{fiorini1998motion}, RVO ($\text{RVO}_\text{c}$)~\cite{van2008reciprocal}, HRVO ($\text{HRVO}_\text{c}$)~\cite{snape2011hybrid}) in the random scenario with different sizes of robots. The bold ones indicate that our approach is superior to the corresponding state of the art.}
\label{tab:performance}
\resizebox{\linewidth}{!}{%
\begin{tabular}{c|cccccc|cccccc|cccccc}
\hline
\multirow{2}{*}{\begin{tabular}[c]{@{}c@{}}Robot Dimen.\\ Ratio \end{tabular}} & \multicolumn{6}{c|}{Completion Rate ($\%$)} & \multicolumn{6}{c|}{Deadlock Rate ($\%$)} & \multicolumn{6}{c}{Average Travel Distance (\si[per-mode=symbol]{\metre}) / std} \\ \cline{2-19} 
& \multicolumn{1}{c}{$\textbf{VO}_\textbf{c}$} & \multicolumn{1}{c}{$\text{VO}_\textbf{p}$} & \multicolumn{1}{c}{$\textbf{RVO}_\textbf{c}$} & \multicolumn{1}{c}{$\text{RVO}_\text{p}$} & \multicolumn{1}{c}{$\textbf{HRVO}_\textbf{c}$} & $\text{HRVO}_\text{p}$ & \multicolumn{1}{c}{$\textbf{VO}_\textbf{c}$} & \multicolumn{1}{c}{$\text{VO}_\textbf{p}$} & \multicolumn{1}{c}{$\textbf{RVO}_\textbf{c}$} & \multicolumn{1}{c}{$\text{RVO}_\text{p}$} & \multicolumn{1}{c}{$\textbf{HRVO}_\textbf{c}$} & $\text{HRVO}_\text{p}$ & \multicolumn{1}{c}{$\textbf{VO}_\textbf{c}$} & \multicolumn{1}{c}{$\text{VO}_\textbf{p}$} & \multicolumn{1}{c}{$\textbf{RVO}_\textbf{c}$} & \multicolumn{1}{c}{$\text{RVO}_\text{p}$} & \multicolumn{1}{c}{$\textbf{HRVO}_\textbf{c}$} & $\text{HRVO}_\text{p}$ \\ \hline
0.4 & \multicolumn{1}{c}{100} & \multicolumn{1}{c}{98} & \multicolumn{1}{c}{100} & \multicolumn{1}{c}{98} & \multicolumn{1}{c}{100} & \textbf{100} & \multicolumn{1}{c}{0} & \multicolumn{1}{c}{\textbf{0}} & \multicolumn{1}{c}{0} & \multicolumn{1}{c}{\textbf{0}} & \multicolumn{1}{c}{0} & \textbf{0} & \multicolumn{1}{c}{7.53/0.40} & \multicolumn{1}{c}{\textbf{7.49/0.39}} & \multicolumn{1}{c}{7.59/0.44} & \multicolumn{1}{c}{\textbf{7.54/0.39}} & \multicolumn{1}{c}{7.51/0.34} & \textbf{7.49/0.38} \\
0.6 & \multicolumn{1}{c}{100} & \multicolumn{1}{c}{\textbf{100}} & \multicolumn{1}{c}{100} & \multicolumn{1}{c}{99} & \multicolumn{1}{c}{100} & \textbf{100} & \multicolumn{1}{c}{0} & \multicolumn{1}{c}{\textbf{0}} & \multicolumn{1}{c}{0} & \multicolumn{1}{c}{\textbf{0}} & \multicolumn{1}{c}{0} & \textbf{0} & \multicolumn{1}{c}{7.72/0.74} & \multicolumn{1}{c}{\textbf{7.64/0.50}} & \multicolumn{1}{c}{7.83/0.67} & \multicolumn{1}{c}{\textbf{7.74/0.59}} & \multicolumn{1}{c}{7.87/0.76} & \textbf{7.64/0.50} \\
0.8 & \multicolumn{1}{c}{100} & \multicolumn{1}{c}{95} & \multicolumn{1}{c}{100} & \multicolumn{1}{c}{96} & \multicolumn{1}{c}{100} & 96 & \multicolumn{1}{c}{0} & \multicolumn{1}{c}{\textbf{0}} & \multicolumn{1}{c}{0} & \multicolumn{1}{c}{\textbf{0}} & \multicolumn{1}{c}{0} & \textbf{0} & \multicolumn{1}{c}{8.04/1.00} & \multicolumn{1}{c}{\textbf{7.91/0.80}} & \multicolumn{1}{c}{8.05/1.04} & \multicolumn{1}{c}{\textbf{7.74/0.56}} & \multicolumn{1}{c}{8.17/1.24} & \textbf{7.76/0.71} \\
1.0 & \multicolumn{1}{c}{85} & \multicolumn{1}{c}{77} & \multicolumn{1}{c}{95} & \multicolumn{1}{c}{\textbf{95}} & \multicolumn{1}{c}{82} & \textbf{94} & \multicolumn{1}{c}{15} & \multicolumn{1}{c}{\textbf{10}} & \multicolumn{1}{c}{5} & \multicolumn{1}{c}{\textbf{5}}  & \multicolumn{1}{c}{13} & \textbf{5} & \multicolumn{1}{c}{8.44/1.19} & \multicolumn{1}{c}{\textbf{8.08/1.09}} & \multicolumn{1}{c}{8.62/1.71} & \multicolumn{1}{c}{\textbf{7.90/0.73}} & \multicolumn{1}{c}{8.67/1.42} & \textbf{7.81/0.54} \\
1.1 & \multicolumn{1}{c}{71} & \multicolumn{1}{c}{\textbf{73}} & \multicolumn{1}{c}{66} & \multicolumn{1}{c}{\textbf{86}} & \multicolumn{1}{c}{63} & \textbf{95} & \multicolumn{1}{c}{15} & \multicolumn{1}{c}{\textbf{10}} & \multicolumn{1}{c}{5} & \multicolumn{1}{c}{\textbf{5}}  & \multicolumn{1}{c}{15} & \textbf{5} & \multicolumn{1}{c}{8.08/1.38} & \multicolumn{1}{c}{\textbf{7.98/0.91}} & \multicolumn{1}{c}{8.09/0.89} & \multicolumn{1}{c}{\textbf{8.02/1.50}} & \multicolumn{1}{c}{8.39/2.09} & \textbf{7.68/0.46} \\
1.2 & \multicolumn{1}{c}{50} & \multicolumn{1}{c}{\textbf{62}} & \multicolumn{1}{c}{33} & \multicolumn{1}{c}{\textbf{69}} & \multicolumn{1}{c}{54} & \textbf{84} & \multicolumn{1}{c}{30} & \multicolumn{1}{c}{\textbf{22}} & \multicolumn{1}{c}{66} & \multicolumn{1}{c}{\textbf{20}}  & \multicolumn{1}{c}{45} & \textbf{11} & \multicolumn{1}{c}{7.53/0.81} & \multicolumn{1}{c}{\textbf{7.27/0.21}} & \multicolumn{1}{c}{7.59/0.97} & \multicolumn{1}{c}{\textbf{7.35/0.19}} & \multicolumn{1}{c}{7.69/0.93} & \textbf{7.29/0.21} \\
1.3 & \multicolumn{1}{c}{42} & \multicolumn{1}{c}{\textbf{47}} & \multicolumn{1}{c}{39} & \multicolumn{1}{c}{\textbf{49}} & \multicolumn{1}{c}{35} & \textbf{59} & \multicolumn{1}{c}{41} & \multicolumn{1}{c}{\textbf{30}} & \multicolumn{1}{c}{59} & \multicolumn{1}{c}{\textbf{36}}  & \multicolumn{1}{c}{64} & \textbf{18} & \multicolumn{1}{c}{7.35/0.21} & \multicolumn{1}{c}{\textbf{7.22/0.10}} & \multicolumn{1}{c}{7.35/0.21} & \multicolumn{1}{c}{\textbf{7.31/0.11}} & \multicolumn{1}{c}{7.35/0.15} & \textbf{7.28/0.17} \\
1.4 & \multicolumn{1}{c}{24} & \multicolumn{1}{c}{\textbf{30}} & \multicolumn{1}{c}{32} & \multicolumn{1}{c}{\textbf{38}} & \multicolumn{1}{c}{25} & \textbf{42} & \multicolumn{1}{c}{56} & \multicolumn{1}{c}{\textbf{55}} & \multicolumn{1}{c}{63} & \multicolumn{1}{c}{\textbf{45}}  & \multicolumn{1}{c}{75} & \textbf{35} & \multicolumn{1}{c}{7.44/0.22} & \multicolumn{1}{c}{\textbf{7.41/0.21}} & \multicolumn{1}{c}{7.47/0.12} & \multicolumn{1}{c}{\textbf{7.44/0.19}} & \multicolumn{1}{c}{7.55/0.35} & \textbf{7.39/0.23} \\
\hline 
\end{tabular}
}
\end{table*}

\subsection{Performance Evaluation with Random Scenarios}
Since each polytope could be encircled by a hyper-ellipse, we could also use the circular-shaped based VO to realize the distributed multi-robot navigation with polytopic shapes.
In each random scenario, we select $8$ rectangle robots with dimensions $1.0 \, \si[per-mode=symbol]{\metre} \times 0.6 \, \si[per-mode=symbol]{\metre}$ as simulation objects, and the initial and goal position of each robot are randomly set on a circle of radius $4 \, \si[per-mode=symbol]{\metre}$ centered at $(5 \, \si[per-mode=symbol]{\metre}, 5 \, \si[per-mode=symbol]{\metre})$.

Three metrics are utilized to evaluate the method's performance: completion rate, deadlock rate, and average travel distance.
The completion rate is the ratio of robots successfully navigating to their goal position without any collisions or deadlock, which describes the method's performance of collision avoidance and navigation.
The deadlock rate is the ratio of cases being stuck somewhere during the navigation without any collisions, which describes the method's performance of navigation.
Furthermore, the average travel distance refers to the average distance traveled by the robot from the initial position to the goal position for all completed cases, which describes the optimality of the method.
All methods ($\text{VO}_\text{p}$, $\text{VO}_\text{c}$, $\text{RVO}_\text{p}$, $\text{RVO}_\text{c}$, $\text{HRVO}_\text{p}$, $\text{HRVO}_\text{c}$) are performed for $100$ trails in the random scenario.
Additionally, we also resize the robots lengths and widths with ratios varying from $0.4$ to $1.4$, e.g., a ratio of $0.4$ means scaling the robot to $0.4$ times of its original size, i.e., $0.4 \, \si[per-mode=symbol]{\metre} \times 0.24 \, \si[per-mode=symbol]{\metre}$.
For these random simulations, we compare the above metrics for different methods and observe how these metrics change with the robot dimension ratio.
We list the results in Tab.~\ref{tab:performance}.
Intuitively, as the size of robot gets bigger, the completion rate decreases, the deadlock rate and the average travel distance increase. 
We notice that when the robot dimension ratio is large ($>$ 1.0), a conservative approximation of robot shape, i.e., considering polytopic-shaped robots as hyper-ellipses, is more likely to result in a deadlock, which results in a low completion rate for $\text{VO}_\text{c}$, $\text{RVO}_\text{c}$ and $\text{HRVO}_\text{c}$.
Additionally, our methods can hold better completion and deadlock rates in this case for navigation tasks.
We also need to note that when the robot dimension ratio is very large (1.4), all methods have a low completion rate due to deadlock, since robots tend to choose a slower velocity in the crowded scenarios.
Moreover, our methods $\text{VO}_\text{p}$, $\text{RVO}_\text{p}$, $\text{HRVO}_\text{p}$ outperform $\text{VO}_\text{c}$, $\text{RVO}_\text{c}$ and $\text{HRVO}_\text{c}$ in terms of the average travel time regardless of the size of robot, which means our methods are more time-optimal.

\begin{remark}
\label{rem:discretization}
We notice that compared to $\text{VO}_\text{c}$, $\text{RVO}_\text{c}$ and $\text{HRVO}_\text{c}$ our methods $\text{VO}_\text{p}$, $\text{RVO}_\text{p}$ and $\text{HRVO}_\text{p}$ obtain slightly lower completion rates when the dimensional ratio is small ($\leq$ 1.0), i.e., the robot's length is less than $1.0 \, \si[per-mode=symbol]{\metre}$, the width is less than $0.6 \, \si[per-mode=symbol]{\metre}$. 
This comes from discretization errors in simulation and can be resolved with angle padding on $\mathbf{vl}$ and $\mathbf{vr}$~\eqref{eq:vl_and_vr}.
We also notice that the average travel distance decreases when the ratio is between $1.0 \sim 1.4$, since only the simple cases of randomized ones could be completed by all methods, leading to a short travel distance in a statistical bias.
\end{remark}

\begin{figure}[h]
    \centering
    \begin{subfigure}{0.49\linewidth}
        \centering
        \includegraphics[width=0.95\linewidth]{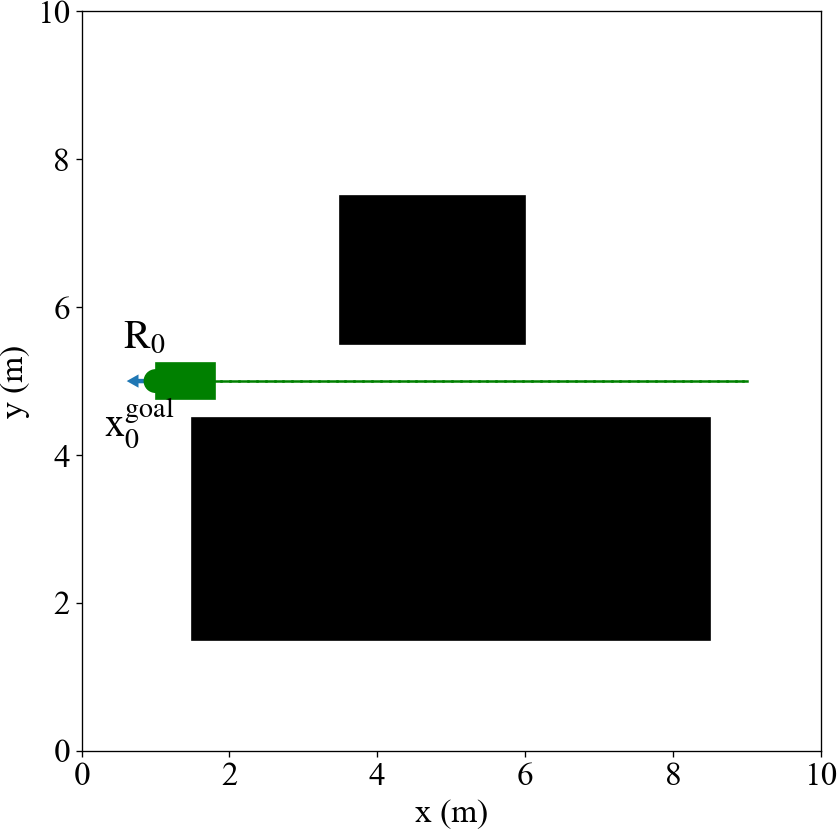}
        \caption{$\text{VO}_\text{p}$}
        \label{subfig:case_compare1}
    \end{subfigure}
    \centering
    \begin{subfigure}{0.49\linewidth}
        \centering
        \includegraphics[width=0.95\linewidth]{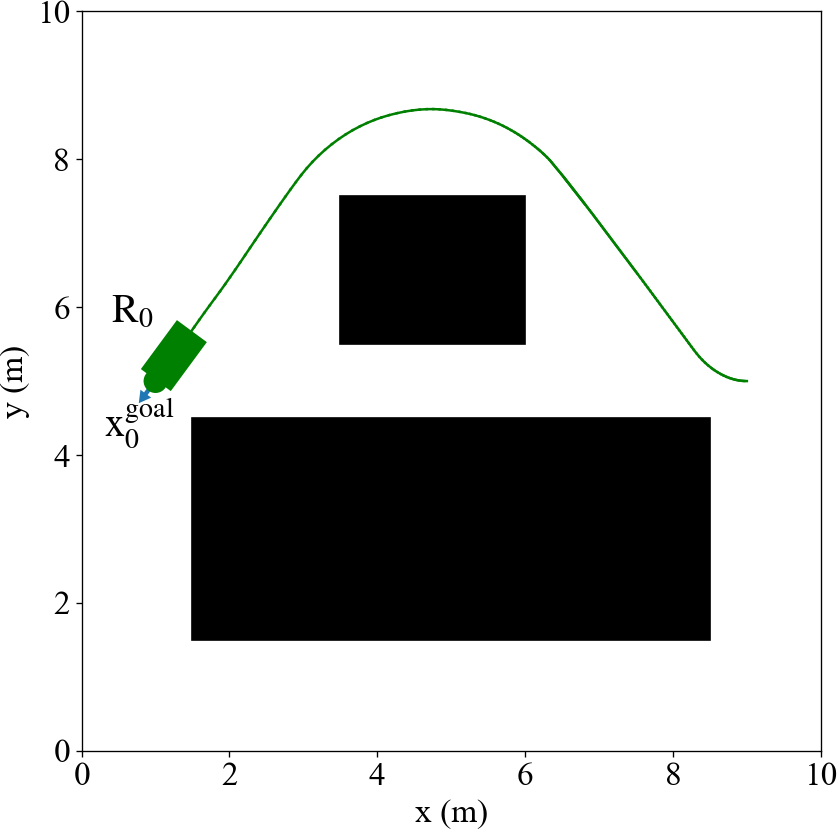}
        \caption{$\text{VO}_\text{c}$}
        \label{subfig:case_compare2}
    \end{subfigure}
    \caption{The comparison of $\text{VO}_\text{p}$ and $\text{VO}_\text{c}$, using $\text{VO}_\text{p}$ could generate a less-conservative trajectory to pass through the corridor between obstacles instead of moving around them.}
    \label{fig:evaluation}
\end{figure}

We also compare $\text{VO}_\text{p}$ and $\text{VO}_\text{c}$ in the turnaround scenario shown in Fig.~\ref{fig:evaluation}, the initial and goal position of the robot is $(9.0 \, \si[per-mode=symbol]{\metre}, 5.0 \, \si[per-mode=symbol]{\metre})$ and $(1.0 \, \si[per-mode=symbol]{\metre}, 5.0 \, \si[per-mode=symbol]{\metre})$, and the shape of the robot is a rectangle of size $0.8 \, \si[per-mode=symbol]{\metre}\times0.5 \, \si[per-mode=symbol]{\metre}$.
Due to conservative estimates, the method based on $\text{VO}_\text{c}$ causes the robot to bypass the obstacle, leading to a longer path, as shown in Fig.~\ref{subfig:case_compare2}, while our approach could guide the robot through the narrow corridor to the goal position directly, as shown in Fig.~\ref{subfig:case_compare2}.
Therefore we verify that our approach is more efficient in travel distance than the circular-shaped based VO.

% Because RVO and HRVO take into account the reactive nature of robots, making them more suitable for the navigation of distributed multi-robot systems, we compare these four types of methods: $\text{RVO}_\text{p}$, $\text{RVO}_\text{c}$, $\text{HRVO}_\text{p}$ and $\text{HRVO}_\text{c}$ in the random scenarios to evaluate the performance of our methods.

Our work also have some shortcomings.
If we select a preferred velocity as in \eqref{eq:v_choose} in a crowded environment with many static obstacles, it will result in deadlock or some unsafe scenarios due to the absence of a global planner in this work, as discussed in Rem.~\ref{rem:global-planner}.
Furthermore, this work doesn't consider model uncertainties, sensor measurement noise, and real world localization and mapping uncertainties.
% For chatgpt: Furthermore, this work doesn't consider model uncertainties, sensor measurement noise, and the uncertainties associated with world localization and mapping, which will be our future works.

\section{Conclusion}
\label{sec:con}
In this paper, we have proposed a velocity obstacle (VO)-based approach for distributed multi-robot navigation with polytopic shapes.
We first proposed an optimization-free approach to realize the collision avoidance between two polytopic objects by constructing the VO between them.
Then, we proposed a VO-based approach for distributed multi-robot navigation with polytopic shapes and validated it in many challenging scenarios.
Numerical simulation results demonstrated that our approach has good navigation performance and outperforms the state of the art in terms of the completion rate, deadlock rate, and average travel distance.
In our future work, we plan to extend our proposed approach to 3D space.

\balance
{
\bibliographystyle{IEEEtran}
\bibliography{references}

% Generated by IEEEtran.bst, version: 1.12 (2007/01/11)
\begin{thebibliography}{10}
\providecommand{\url}[1]{#1}
\csname url@samestyle\endcsname
\providecommand{\newblock}{\relax}
\providecommand{\bibinfo}[2]{#2}
\providecommand{\BIBentrySTDinterwordspacing}{\spaceskip=0pt\relax}
\providecommand{\BIBentryALTinterwordstretchfactor}{4}
\providecommand{\BIBentryALTinterwordspacing}{\spaceskip=\fontdimen2\font plus
\BIBentryALTinterwordstretchfactor\fontdimen3\font minus \fontdimen4\font\relax}
\providecommand{\BIBforeignlanguage}[2]{{%
\expandafter\ifx\csname l@#1\endcsname\relax
\typeout{** WARNING: IEEEtran.bst: No hyphenation pattern has been}%
\typeout{** loaded for the language `#1'. Using the pattern for}%
\typeout{** the default language instead.}%
\else
\language=\csname l@#1\endcsname
\fi
#2}}
\providecommand{\BIBdecl}{\relax}
\BIBdecl

\bibitem{chung2018survey}
S.-J. Chung, A.~A. Paranjape, P.~Dames, S.~Shen, and V.~Kumar, ``A survey on aerial swarm robotics,'' \emph{IEEE Transactions on Robotics}, vol.~34, no.~4, pp. 837--855, 2018.

\bibitem{pian2022distributed}
P.~Yu and D.~V. Dimarogonas, ``Distributed motion coordination for multirobot systems under ltl specifications,'' \emph{IEEE Transactions on Robotics}, vol.~38, no.~2, pp. 1047--1062, 2022.

\bibitem{queralta2020collaborative}
J.~P. Queralta, J.~Taipalmaa, B.~Can~Pullinen, V.~K. Sarker, T.~Nguyen~Gia, H.~Tenhunen, M.~Gabbouj, J.~Raitoharju, and T.~Westerlund, ``Collaborative multi-robot search and rescue: Planning, coordination, perception, and active vision,'' \emph{IEEE Access}, vol.~8, pp. 191\,617--191\,643, 2020.

\bibitem{fiorini1998motion}
P.~Fiorini and Z.~Shiller, ``Motion planning in dynamic environments using velocity obstacles,'' \emph{The international journal of robotics research}, vol.~17, no.~7, pp. 760--772, 1998.

\bibitem{ziegler2010fast}
J.~Ziegler and C.~Stiller, ``Fast collision checking for intelligent vehicle motion planning,'' in \emph{2010 IEEE Intelligent Vehicles Symposium}, 2010, pp. 518--522.

\bibitem{grossmann2002review}
I.~E. Grossmann, ``Review of nonlinear mixed-integer and disjunctive programming techniques,'' \emph{Optimization and engineering}, vol.~3, no.~3, pp. 227--252, 2002.

\bibitem{li2015unified}
B.~Li and Z.~Shao, ``A unified motion planning method for parking an autonomous vehicle in the presence of irregularly placed obstacles,'' \emph{Knowledge-Based Systems}, vol.~86, pp. 11--20, 2015.

\bibitem{deits2015efficient}
R.~Deits and R.~Tedrake, ``Efficient mixed-integer planning for uavs in cluttered environments,'' in \emph{2015 IEEE International Conference on Robotics and Automation (ICRA)}, 2015, pp. 42--49.

\bibitem{zhang2018autonomous}
X.~Zhang, A.~Liniger, A.~Sakai, and F.~Borrelli, ``Autonomous parking using optimization-based collision avoidance,'' in \emph{2018 IEEE Conference on Decision and Control (CDC)}, 2018, pp. 4327--4332.

\bibitem{zhang2020optimization}
X.~Zhang, A.~Liniger, and F.~Borrelli, ``Optimization-based collision avoidance,'' \emph{IEEE Transactions on Control Systems Technology}, vol.~29, no.~3, pp. 972--983, 2021.

\bibitem{firoozi2020distributed}
R.~Firoozi, L.~Ferranti, X.~Zhang, S.~Nejadnik, and F.~Borrelli, ``A distributed multi-robot coordination algorithm for navigation in tight environments,'' \emph{arXiv preprint arXiv:2006.11492}, 2020.

\bibitem{thirugnanam2021duality}
A.~Thirugnanam, J.~Zeng, and K.~Sreenath, ``Duality-based convex optimization for real-time obstacle avoidance between polytopes with control barrier functions,'' in \emph{2022 American Control Conference (ACC)}, 2022, pp. 2239--2246.

\bibitem{thirugnanam2022safety}
{Thirugnanam, Akshay and Zeng, Jun and Sreenath, Koushil}, ``Safety-critical control and planning for obstacle avoidance between polytopes with control barrier functions,'' in \emph{2022 International Conference on Robotics and Automation (ICRA)}, 2022, pp. 286--292.

\bibitem{glotfelter2017nonsmooth}
P.~Glotfelter, J.~Cort{\'e}s, and M.~Egerstedt, ``Nonsmooth barrier functions with applications to multi-robot systems,'' \emph{IEEE control systems letters}, vol.~1, no.~2, pp. 310--315, 2017.

\bibitem{zeng2021safety}
J.~Zeng, B.~Zhang, and K.~Sreenath, ``Safety-critical model predictive control with discrete-time control barrier function,'' in \emph{2021 American Control Conference (ACC)}, 2021, pp. 3882--3889.

\bibitem{asiain2020navigation}
J.~Asiain and J.~Godoy, ``Navigation in large groups of robots,'' \emph{Current Robotics Reports}, vol.~1, no.~4, pp. 203--213, 2020.

\bibitem{raibail2022decentralized}
M.~Raibail, A.~H.~A. Rahman, G.~J. AL-Anizy, M.~F. Nasrudin, M.~S.~M. Nadzir, N.~M.~R. Noraini, and T.~S. Yee, ``Decentralized multi-robot collision avoidance: A systematic review from 2015 to 2021,'' \emph{Symmetry}, vol.~14, no.~3, p. 610, 2022.

\bibitem{van2008reciprocal}
J.~van~den Berg, M.~Lin, and D.~Manocha, ``Reciprocal velocity obstacles for real-time multi-agent navigation,'' in \emph{2008 IEEE International Conference on Robotics and Automation}, 2008, pp. 1928--1935.

\bibitem{berg2011reciprocal}
J.~v.~d. Berg, S.~J. Guy, M.~Lin, and D.~Manocha, ``Reciprocal n-body collision avoidance,'' in \emph{Robotics research}.\hskip 1em plus 0.5em minus 0.4em\relax Springer Berlin Heidelberg, 2011, pp. 3--19.

\bibitem{snape2011hybrid}
J.~Snape, J.~v.~d. Berg, S.~J. Guy, and D.~Manocha, ``The hybrid reciprocal velocity obstacle,'' \emph{IEEE Transactions on Robotics}, vol.~27, no.~4, pp. 696--706, 2011.

\bibitem{alonso2012reciprocal}
J.~Alonso-Mora, A.~Breitenmoser, P.~Beardsley, and R.~Siegwart, ``Reciprocal collision avoidance for multiple car-like robots,'' in \emph{2012 IEEE International Conference on Robotics and Automation}, 2012, pp. 360--366.

\bibitem{alonso2013optimal}
J.~Alonso-Mora, A.~Breitenmoser, M.~Rufli, P.~Beardsley, and R.~Siegwart, ``Optimal reciprocal collision avoidance for multiple non-holonomic robots,'' in \emph{Distributed autonomous robotic systems}.\hskip 1em plus 0.5em minus 0.4em\relax Springer, 2013, pp. 203--216.

\bibitem{long2018towards}
P.~Long, T.~Fan, X.~Liao, W.~Liu, H.~Zhang, and J.~Pan, ``Towards optimally decentralized multi-robot collision avoidance via deep reinforcement learning,'' in \emph{2018 IEEE International Conference on Robotics and Automation (ICRA)}, 2018, pp. 6252--6259.

\bibitem{li2019reciprocal}
H.~Li, B.~Weng, A.~Gupta, J.~Pan, and W.~Zhang, ``Reciprocal collision avoidance for general nonlinear agents using reinforcement learning,'' \emph{arXiv preprint arXiv:1910.10887}, 2019.

\bibitem{han2022reinforcement}
R.~Han, S.~Chen, S.~Wang, Z.~Zhang, R.~Gao, Q.~Hao, and J.~Pan, ``Reinforcement learned distributed multi-robot navigation with reciprocal velocity obstacle shaped rewards,'' \emph{IEEE Robotics and Automation Letters}, vol.~7, no.~3, pp. 5896--5903, 2022.

\bibitem{karaman2010optimal}
S.~Karaman and E.~Frazzoli, ``Optimal kinodynamic motion planning using incremental sampling-based methods,'' in \emph{49th IEEE Conference on Decision and Control (CDC)}, 2010, pp. 7681--7687.

\end{thebibliography}
}

\end{document}